%% file: PaperForReview.tex
\documentclass[10pt,twocolumn,letterpaper]{article}

\usepackage{cvpr}              

\usepackage{graphicx}
\usepackage{amsmath}
\usepackage{amssymb}
\usepackage{booktabs,caption}
\usepackage{multirow}
\usepackage{bm}
\usepackage{dsfont}
\usepackage[linesnumbered,ruled,vlined]{algorithm2e}
\usepackage{makecell}
\usepackage[flushleft]{threeparttable}
\usepackage{enumitem}
\usepackage[accsupp]{axessibility} 
\usepackage{siunitx}
\usepackage[dvipsnames]{xcolor}

\graphicspath{{figures/}}
\definecolor{cvprblue}{rgb}{0.21,0.49,0.74}
\usepackage[pagebackref,breaklinks,colorlinks,citecolor=cvprblue]{hyperref}

\def\eg{\emph{e.g.\ }}
\def\ie{\emph{i.e.\ }}

\usepackage[capitalize]{cleveref}
\crefname{section}{Sec.}{Secs.}
\Crefname{section}{Section}{Sections}
\Crefname{table}{Table}{Tables}
\crefname{table}{Tab.}{Tabs.}

\def\confName{CVPR}
\def\confYear{2024}

\begin{document}

\title{NTIRE 2024 Challenge on Low Light Image Enhancement: Methods and Results}

\author{Xiaoning Liu$^*$ \and
Zongwei Wu$^*$ \and
Ao Li$^*$ \and
Florin-Alexandru Vasluianu$^*$ \and
Yulun Zhang$^*$ \and
Shuhang Gu$^*$ \and
Le Zhang$^*$ \and
Ce Zhu$^*$ \and
Radu Timofte$^*$ \and
Zhi Jin \and
Hongjun Wu \and
Chenxi Wang \and
Haitao Ling \and
Yuanhao Cai \and
Hao Bian \and
Yuxin Zheng \and
Jing Lin \and
Alan Yuille \and
Ben Shao \and 
Jin Guo \and
Tianli Liu \and
Mohao Wu \and
Yixu Feng \and
Shuo Hou \and
Haotian Lin \and
Yu Zhu \and
Peng Wu \and
Wei Dong \and
Jinqiu Sun \and
Yanning Zhang \and
Qingsen Yan \and
Wenbin Zou \and
Weipeng Yang \and
Yunxiang Li \and 
Qiaomu Wei \and
Tian Ye \and
Sixiang Chen \and
Zhao Zhang \and
Suiyi Zhao \and 
Bo Wang \and
Yan Luo \and
Zhichao Zuo \and
Mingshen Wang \and
Junhu Wang \and
Yanyan Wei \and
Xiaopeng Sun \and
Yu Gao \and 
Jiancheng Huang \and
Hongming Chen \and 
Xiang Chen \and
Hui Tang \and
Yuanbin Chen \and
Yuanbo Zhou \and
Xinwei Dai \and
Xintao Qiu \and
Wei Deng \and
Qinquan Gao \and
Tong Tong \and
Mingjia Li \and
Jin Hu\and
Xinyu He\and
Xiaojie Guo\and
Sabarinathan \and
K Uma \and
A Sasithradevi \and
B Sathya Bama \and
S. Mohamed Mansoor Roomi \and
V.Srivatsav \and
Jinjuan Wang \and 
Long Sun \and 
Qiuying Chen\and 
Jiahong Shao\and 
Yizhi Zhang \and
Marcos V. Conde \and
Daniel Feijoo \and
Juan C. Benito \and 
Alvaro Garc\'{i}a \and
Jaeho Lee\and
Seongwan Kim \and
Sharif S M A \and 
Nodirkhuja Khujaev\and 
Roman Tsoy \and
Ali Murtaza \and
Uswah Khairuddin \and
Ahmad 'Athif Mohd Faudzi\and
Sampada Malagi \and
Amogh Joshi \and
Nikhil Akalwadi \and
Chaitra Desai \and
Ramesh Ashok Tabib \and
Uma Mudenagudi \and
Wenyi Lian \and
Wenjing Lian \and
Jagadeesh Kalyanshetti \and
Vijayalaxmi Ashok Aralikatti \and
Palani Yashaswini \and
Nitish Upasi \and
Dikshit Hegde \and
Ujwala Patil \and
Sujata C \and
Xingzhuo Yan \and
Wei Hao \and
Minghan Fu \and
Pooja choksy \and
Anjali Sarvaiya  \and
Kishor Upla \and
Kiran Raja \and
Hailong Yan \and
Yunkai Zhang \and
Baiang Li \and
Jingyi Zhang  \and
Huan Zheng \and
}
\maketitle
\let\thefootnote\relax\footnotetext{$^*$ X. Liu, Z. Wu, A. Li, F. Vasluianu, Y. Zhang, S. Gu, L. Zhang, C. Zhu and R. Timofte were the challenge organizers, while the other authors participated in the challenge. Each team described their own method in the report. Appendix~\ref{sec:teams} contains the authors' teams and affiliations. \\ NTIRE 2024 webpage:~\url{https://cvlai.net/ntire/2024}.\\
Code:~\url{https://github.com/AVC2-UESTC/NTIRE24-LLE/}} 

\begin{abstract}
This paper reviews the NTIRE 2024 low light image enhancement challenge, highlighting the proposed solutions and results. The aim of this challenge is to discover an effective network design or solution capable of generating brighter, clearer, and visually appealing results when dealing with a variety of conditions, including ultra-high resolution (4K and beyond), non-uniform illumination, backlighting, extreme darkness, and night scenes. A notable total of 428 participants registered for the challenge, with 22 teams ultimately making valid submissions. This paper meticulously evaluates the state-of-the-art advancements in enhancing low-light images, reflecting the significant progress and creativity in this field.
\end{abstract}

\section{Introduction}
\label{sec:introduction}
Low light Image enhancement (LLIE), a pivotal yet challenging task in computer vision, aims to improve visibility and contrast across a broad spectrum of low-light scenarios, including uneven illumination, extreme darkness, backlighting, and night. Additionally, LLIE strives to correct imperfections like noise, artifacts, and color distortion. These challenges, arising in darkness or through illumination enhancement, affect both human visual perception and downstream tasks like object detection and scene segmentation.

As deep learning technology improves by leaps and bounds, remarkable advances in LLIE are impressive. However, current cutting-edge methods~\cite{guo2020zero,cai2023retinexformer,jiang2023low,xu2022snr,ma2022toward,liu2021retinex,zamir2020learning} not only struggle to adapt to complex and variable low-light conditions but also face major challenges when being deployed on consumer-grade devices such as smartphones and cameras. This is primarily constrained by the limitations of current datasets \cite{wei2018deep,bychkovsky2011learning} that suffer from limited scene diversity (notably insufficient night scenes), low resolution, and overly simplistic lighting conditions. Additionally, the high complexity of models \cite{cai2023retinexformer,hou2024global,yang2023implicit,wang2023ultra,jiang2023low} hampers their ability to handle ultra-high resolution images that are commonly captured by smartphones.

To address the aforementioned challenges, we launched the inaugural Low Light Enhancement Challenge at the 2024 New Trends in Image Restoration and Enhancement (NTIRE 2024) workshop. The objective of this challenge is
to foster innovative thinking and discover solutions that significantly improve image quality under various low-light conditions. To this end, we have built an LLIE dataset that features a wide range of scenes, encompassing various low-light conditions. These images include indoor and outdoor locations under both daylight and nighttime conditions, with the challenge focusing on enhancing image quality across these diverse settings.

In conclusion, this challenge aims to set a new benchmark for LLIE while highlighting specific challenges and research questions in this domain. We hope that it will inspire the research community to explore these pressing issues and identify emerging trends. Moreover, this challenge is one of the NTIRE 2024 Workshop associated challenges on: dense and non-homogeneous dehazing~\cite{ntire2024dehazing}, night photography rendering~\cite{ntire2024night}, blind compressed image enhancement~\cite{ntire2024compressed}, shadow removal~\cite{ntire2024shadow}, efficient super resolution~\cite{ntire2024efficientsr}, image super resolution ($\times$4)~\cite{ntire2024srx4}, light field image super-resolution~\cite{ntire2024lightfield}, stereo image super-resolution~\cite{ntire2024stereosr}, HR depth from images of specular and transparent surfaces~\cite{ntire2024depth}, bracketing image restoration and enhancement~\cite{ntire2024bracketing}, portrait quality assessment~\cite{ntire2024QA_portrait}, quality assessment for AI-generated content~\cite{ntire2024QA_AI}, restore any image model (RAIM) in the wild~\cite{ntire2024raim}, RAW image super-resolution~\cite{ntire2024rawsr}, short-form UGC video quality assessment~\cite{ntire2024QA_UGC}, low light enhancement~\cite{ntire2024lowlight}, and RAW burst alignment and ISP challenge.

\section{NTIRE 2024 Low Light Enhancement Challenge}
The objectives of this challenge are threefold: (1) to advance research in the field of low light enhancement, (2) to facilitate systematic comparisons among different methodologies, and (3) to provide a forum for both academic and industrial stakeholders to interact, deliberate, and potentially forge partnerships. This section delves into the detailed aspects of the challenge.

\subsection{Dataset}
Although some low light datasets, such as~LOL \cite{wei2018deep} and MIT-Adobe FiveK \cite{bychkovsky2011learning}, have been widely applied in the field, they exhibit limitations previously mentioned, including limited resolution, monotonous scene content, and uniform illumination levels. These factors often restrict the generalization capabilities of cutting-edge models \cite{cai2023retinexformer, jiang2023low, xu2022snr, ma2022toward, liu2021retinex, zamir2020learning}, posing significant challenges in adapting to diverse low-light conditions, especially when implemented on consumer-grade devices like smartphones and cameras. To address these issues, this challenge introduces a rich array of contest scenarios, covering a variety of lighting conditions such as dim environments, extreme darkness, non-uniform illumination, backlighting, and night scenes, applicable to both indoor and outdoor settings during day and night, with image resolutions up to 4K and beyond. Specifically, the dataset includes 230 training scenes, along with 35 validation and 35 testing scenes. The ground truth (GT) images for both validation and testing were kept concealed from the participants throughout the challenge. Further details about the dataset will be provided in subsequent works.

\subsection{Tracks and Competition}
\noindent\textbf{Ranking statistic.~}In this challenge, we primarily use peak signal-to-noise ratio (PSNR), structural similarity (SSIM), and Learned Perceptual Image Patch Similarity (LPIPS) \cite{zhang2018unreasonable} as the criteria for comparing models submitted by participants. Given that LPIPS is a learned perceptual evaluation metric and our competition focuses mainly on quantitative assessments, We use LPIPS as a supplementary evaluation tool when the quantitative evaluations of two methods are indistinguishable. As listed in~\cref{tbl:ntire24_results}, 
``Final Rank'' represents a composite metric, calculated through a weighted sum of PSNR (60\%) and SSIM (40\%). This way offers a comprehensive evaluation of the effectiveness of each solution in enhancing low-light conditions.

\noindent\textbf{Challenge phases.~}
\textit{(1) Development and validation phase}: Participants were granted access to 230 training image pairs and 35 validation inputs from our constructed dataset. It is noteworthy that the GT images for the validation set were hidden from the participants. Participants had the opportunity to submit their enhanced results to the evaluation server, which calculated the PSNR and SSIM for the enhanced images generated by their models and provided immediate feedback.
\textit{(2) Testing phase:~}Participants gained access to 35 testing low light images from the built dataset, with the ground-truth images remaining undisclosed. Submissions of their enhanced outputs were made to the Codalab evaluation server, accompanied by an email to the organizers containing the code and a factsheet. The organizers subsequently verified and executed the provided code to derive the final results, which were shared with participants upon the conclusion of the challenge.
\input{final_results}

\section{Challenge Results and Discussion}
The results of the low light enhancement challenge are detailed in~\cref{tbl:ntire24_results}, which evaluates and ranks the performances of 22 teams. Notably, one team, AiRiA\_Vision, voluntarily withdrew from the ranking due to issues related to their model design. The evaluation leverages two key performance metrics: PSNR and SSIM. The metrics are calculated based on a test set comprising 35 inputs from the built dataset, thereby ensuring the challenge's integrity and mitigating the risk of overfitting to the validation set.

The top-ranked teams in the challenge boast higher PSNR and SSIM values, signifying superior performance, while the lower-ranked teams exhibit lower values, indicative of suboptimal performance. Notably, two teams achieved a PSNR of over 25 dB, meeting our prior expectations for this metric. For more detailed information on the low-light enhancement methods employed, please refer to~\cref{sec:methods_and_teams}, which discusses the specific solutions provided by each team.

Due to the ultra-high resolution (4K and beyond) of the input images, many teams opted for multi-scale strategies to implement enhancement. While they has indeed reduced computational consumption to some extent, currently, almost all models are unable to perform inference on a plain GPU,~\eg, one with 12G memory. For the next competition, we will consider including NIQE \cite{mittal2012making} along with metrics measuring model efficiency, such as inference time, model parameter count, computational complexity (FLOPs), and memory consumption. This will promote the application of ultra-high-resolution low light enhancement on smartphones.

\section{Challenge Methods and Teams}
\label{sec:methods_and_teams}

\input{team01_SYSU_FVL_T2/main}
\input{team02_Retinexformer/main}
\input{team03_DH_AISP/main}
\input{team04_NWPU_DiffLight/main}

\input{team05_GiantPandaCV/main}

\input{team06_LVGroup_HFUT/main}

\input{team07_Try1try8/main}
\input{team08_Pixel_warrior/main}
\input{team09_HuiT/main}
\input{team10_X_LIME/main}
\input{team11_Image_Lab/main}
\input{team12_dgzzqteam/main}

\input{team13_Cidaut_AI/main}
\input{team14_OptDev/main}
\input{team15_ataza/main}
\input{team16_KLETech_CEVI_LowlightHypnotise/main}
\input{team17_221B/main}
\input{team18_KLETech_CEVI_Dark_Knights/main}
\input{team19_BFU_LL/main}

\input{team20_SVNIT_NTNU/main}
\input{team21_yanhailong/main}

\input{team22_Mishka/main}

\section*{Acknowledgements}
This work was partially supported by the Humboldt Foundation. We thank the NTIRE 2024 sponsors: Meta Reality Labs, OPPO, KuaiShou, Huawei and University of W\"urzburg (Computer Vision Lab).

\appendix

\section{Teams and affiliations}
\label{sec:teams}

\subsection*{NTIRE 2024 team}
\noindent\textit{\textbf{Title: }} NTIRE 2024 Low Light Enhancement Challenge\\
\noindent\textit{\textbf{Members: }} \\
Xiaoning Liu$^1$ (\href{mailto:liuxiaoning2016@sina.com}{liuxiaoning2016@sina.com}),\\
Zongwei Wu$^2$
(\href{mailto:zongwei.wu@uni-wuerzburg.de}{zongwei.wu@uni-wuerzburg.de}),\\
Ao Li$^1$
(\href{mailto:aoli@std.uestc.edu.cn}{aoli@std.uestc.edu.cn}),\\
Florin Vasluianu$^2$
(\href{mailto:florin-alexandru.vasluianu@uni-wuerzburg.de}{florin-alexandru.vasluianu@uni-wuerzburg.de}),\\
Yulun Zhang$^3$ (\href{mailto:yulzhang@ethz.ch}{yulun100@gmail.com}),\\
Shuhang Gu$^1$ (\href{mailto:shuhanggu@gmail.com}{shuhanggu@gmail.com}),\\
Le Zhang$^1$
(\href{mailto:lezhang@uestc.edu.cn}{lezhang@uestc.edu.cn}),\\
Ce Zhu$^1$
(\href{mailto:eczhu@uestc.edu.cn}{eczhu@uestc.edu.cn}),\\
Radu Timofte$^{2,4}$ (\href{mailto:radu.timofte@uni-wuerzburg.de}{radu.timofte@uni-wuerzburg.de})\\
\noindent\textit{\textbf{Affiliations: }}\\
$^1$ University of Electronic Science and Technology of China, China\\
$^2$ Computer Vision Lab, University of W\"urzburg, Germany\\
$^3$ Shanghai Jiao Tong University, China\\
$^4$ Computer Vision Lab, ETH Zurich, Switzerland\\

\input{team01_SYSU_FVL_T2/affiliation}
\input{team02_Retinexformer/affiliation}
\input{team03_DH_AISP/affiliation}

\input{team04_NWPU_DiffLight/affiliation}
\input{team05_GiantPandaCV/affiliation}
\input{team06_LVGroup_HFUT/affiliation}
\input{team07_Try1try8/affiliation}
\input{team08_Pixel_warrior/affiliation}
\input{team09_HuiT/affiliation}
\input{team10_X_LIME/affiliation}
\input{team11_Image_Lab/affiliation}
\input{team12_dgzzqteam/affiliation}
\input{team13_Cidaut_AI/affiliation}
\input{team14_OptDev/affiliation}
\input{team15_ataza/affiliation}

\input{team16_KLETech_CEVI_LowlightHypnotise/affiliation}

\input{team17_221B/affiliation}
\input{team18_KLETech_CEVI_Dark_Knights/affiliation}

\input{team19_BFU_LL/affiliation}
\input{team20_SVNIT_NTNU/affiliation}

\input{team21_yanhailong/affiliation}
\input{team22_Mishka/affiliation}

{\small
\bibliographystyle{ieee_fullname}
\bibliography{egbib}
}

\end{document}

%% file: final_results.tex
\begin{table*}[!t]
\footnotesize
\centering
\caption{Evaluation and Rankings in the NTIRE 2024 Low Light Enhancement Challenge. This table presents a comprehensive comparison of participant solutions across multiple metrics: PSNR, SSIM, and LPIPS. ``Rank PSNR" and ``Rank SSIM" indicate the respective standings of participants based on their performance in PSNR and SSIM metrics on the challenge's test dataset. ``Final Rank" represents a composite metric, derived from a weighted sum of PSNR (60\%) and SSIM (40\%), which provides an overall assessment of each solution's effectiveness in low light enhancement.}
\begin{tabular}{ccccccc}
\toprule
\textbf{Team}                & \textbf{PSNR} & \textbf{SSIM} & \textbf{LPIPS} & \textbf{Rank PSNR} & \textbf{Rank SSIM} & \textbf{Final Rank} \\ 
\midrule
SYSU-FVL-T2                  & 25.52         & 0.8637        & 0.1221         & 1                  & 1                  & 1                   \\
Retinexformer \cite{cai2023retinexformer}                & 25.30         & 0.8525        & 0.1424         & 2                  & 4                  & 2                   \\
DH-AISP                      & 24.97         & 0.8528        & 0.1235         & 3                  & 3                  & 3                   \\
NWPU-DiffLight               & 24.78         & 0.8556        & 0.1673         & 6                  & 2                  & 4                   \\
GiantPandaCV                 & 24.83         & 0.8474        & 0.1353         & 5                  & 7                  & 5                   \\
LVGroup\_HFUT                & 24.88         & 0.8395        & 0.1371         & 4                  & 10                  & 6                  \\
Try1try8                     & 24.49         & 0.8483        & 0.1359         & 8                  & 6                  & 7                   \\
Pixel\_warrior               & 24.74         & 0.8416        & 0.1514         & 7                  & 9                  & 8                   \\
HuiT                         & 24.13         & 0.8484        & 0.1436         & 10                 & 5                  & 9                   \\
X-LIME                       & 24.28         & 0.8446        & 0.1298         & 9                  & 8                  & 10                  \\
Image Lab                    & 23.63         & 0.8235        & 0.1673         & 11                 & 12                 & 11                  \\
dgzzqteam                    & 23.28         & 0.8385        & 0.1406         & 12                 & 11                 & 12                  \\
Cidaut AI (InstructIR~\cite{conde2024high})                   & 23.07         & 0.8075        & 0.1559         & 13                 & 16                 & 13                  \\
OptDev                       & 22.93         & 0.8097        & 0.1592         & 14                 & 15                 & 14                  \\
ataza                        & 22.51         & 0.8161        & 0.1404         & 18                 & 13                 & 15                  \\
KLETech-CEVI\_LowlightHypnotise & 22.85         & 0.7828        & 0.1823         & 15               & 18                 & 16                  \\
221B                         & 22.04         & 0.8141        & 0.1084         & 19                 & 14                 & 17                  \\
KLETech-CEVI\_Dark\_Knights   & 22.76         & 0.7843        & 0.1806         & 17                 & 17                 & 18                  \\
BFU-LL                       & 22.78         & 0.7792        & 0.1826         & 16                 & 19                 & 19                  \\
SVNIT\_NTNU                  & 20.32         & 0.7718        & 0.3089         & 20                 & 20                 & 20                  \\
yanhailong                   & 20.07         & 0.6881        & 0.3133         & 21                 & 22                 & 21                  \\
Mishka                       & 18.19         & 0.7161        & 0.2712         & 22                 & 21                 & 22                  \\
\bottomrule
\end{tabular}
\label{tbl:ntire24_results}
\end{table*}

%% file: team01_SYSU_FVL_T2/main.tex
\subsection{SYSU-FVL-T2}
\begin{figure}[!ht]
  \centering
 \includegraphics[width=\linewidth]{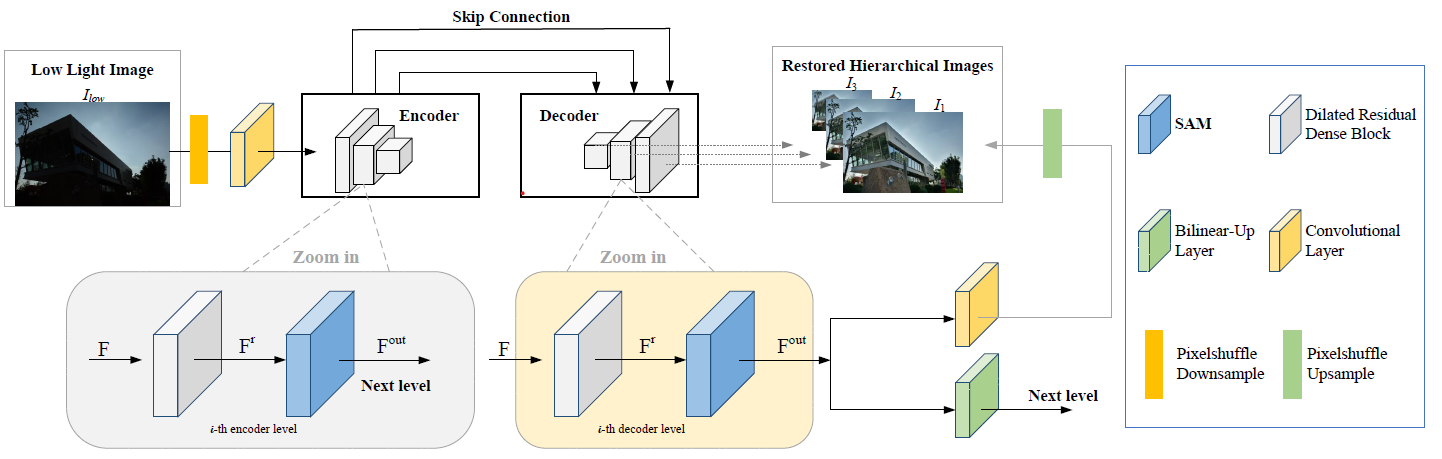}
\caption{The backbone network employed in our method. (Reproduced from ESDNet \cite{uhdm})}
\label{fig_esdnet}%
\end{figure}
\noindent\textbf{Description:}
As shown in~\cref{fig_esdnet}, ESDNet-L \cite{uhdm} is employed as the backbone in our proposed method for low-light image enhancement. The backbone mainly consists of an encoder-decoder network in three feature scales with skip-connections. Different scales of features are generated by adopting the bilinear interpolation. In each scale, the Semantic-Aligned Scale-Aware Modules (SAM) are stacked to enhance the ability of the model to address scale variations. SAM incorporates a pyramid context extraction module and a cross-scale dynamic fusion module to selectively fuse multi-scale features.

Furthermore, for network training, the loss function $L_{total}$ is set for outputs in three scales $\hat{I}_1, \hat{I}_2, \hat{I}_3$:
\begin{equation}
L_{total}=\sum_{i=1}^3{L_C\left( I_i,\hat{I}_i \right)}+\lambda \cdot L_P\left( I_i,\hat{I}_i \right),    
\end{equation}
where $\lambda$ is set as 0.04, ${I}_i$ stands for ground truth image, $\hat{I}_i$ stands for enhanced image, $L_C$ stands for Charbonnier loss \cite{chan}, and $L_P$ stands for perceptual loss using pretrained VGG19 model \cite{vgg19}.

\noindent\textbf{Implementation:}
The proposed method is based on Python 3.8 and the experiment is conducted on one NVIDIA RTX A6000 GPU (49G). Inspired by MIRNet-v2 \cite{mirv2}, we adopt a progressive training strategy. The model is trained for 150000 iterations from scratch and optimized by Adam \cite{adam}. During the training process, at the first stage, the batch size is set to 8, and we randomly crop square image patches of size 720 for training. Subsequently, after 46000, 32000, 24000, 18000, and 18000 iterations, the batch size decreases to 4, 4, 2, 2, and 1, respectively. The lengths of the square image patches are set to 1024, 1024, 1280, 1280, and 1600 for the respective stages. The learning rate is set as $2\times10^{-4}$ initially and is scheduled by cyclic cosine annealing \cite{sgdr}. During the testing phase, the whole low-light image is fed into the network, and the enhanced image is obtained directly. The batchsize in 
 testing is set as 1.

%% file: team02_Retinexformer/main.tex
\subsection{Retinexformer}
The proposed code, pre-trained models, results, and training logs are all publicly available at \url{https://github.com/caiyuanhao1998/Retinexformer}.
\begin{figure*}[t]
	\begin{center}
		\begin{tabular}[t]{c} \hspace{-3mm}
	\includegraphics[width=0.95\linewidth]{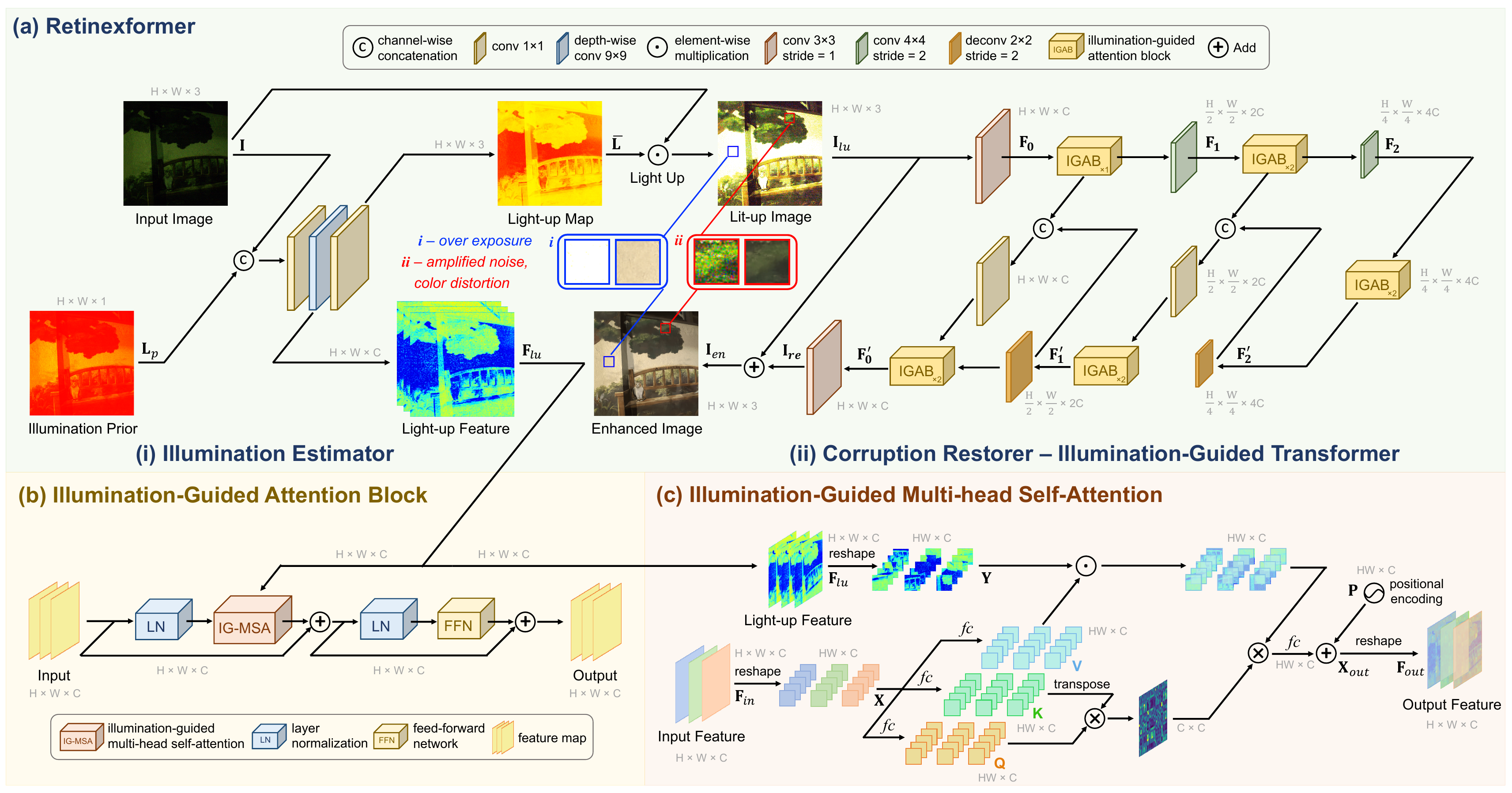}
		\end{tabular}
	\end{center}
	\vspace*{-2.5mm}
	\caption{\small The overview of Retinexformer~\cite{cai2023retinexformer}. (a) Retinexformer adopts the proposed ORF that consists of an illumination estimator (i) and a corruption restorer (ii)  IGT. (b) The basic unit of IGT is IGAB, which is composed of two layer normalization (LN), an IG-MSA and a feed-forward network (FFN). (c) IG-MSA uses the illumination representations captured by ORF  to direct the computation of self-attention.}
	\label{fig:pipeline}
\end{figure*}

\vspace{1mm}
\noindent\textbf{Description:} The team directly adopt their ICCV 2023 work Retinexformer~\cite{cai2023retinexformer} to participate this challenge. \cref{fig:pipeline} illustrates the overall architecture of their method. As shown in~\cref{fig:pipeline} (a), Retinexformer is based on their formulated One-stage Retinex-based Framework (ORF). Their Retinexformer takes a low-light image $\mathbf{I} \in \mathbb{R}^{H\times W\times 3}$ as input and reconstruct its enhanced counterpart $\mathbf{I}_{en}  \in \mathbb{R}^{H\times W\times 3}$. The original Retinex model assumes that the low-light image $\mathbf{I}$ is corruption-free and can be decomposed into a reflectance image  $\mathbf{R} \in \mathbb{R}^{H\times W\times 3}$ and an illumination map  $\mathbf{L} \in \mathbb{R}^{H\times W}$ as:
\begin{equation}
\mathbf{I} = \mathbf{R} \odot \mathbf{L},
\label{eq:retinex_ori}
\end{equation}
where $\odot$ denotes the element-wise multiplication. However, this corruption-free assumption is consistent with the real under-exposed scenes, where corruptions are inevitably introduced by the high-ISO and long-exposure imaging settings, as well as the light-up process. To model the corruptions, they introduce perturbation terms for $\mathbf{R}$ and $\mathbf{L}$ and reformulate~\cref{eq:retinex_ori} as:
\begin{equation}
\begin{aligned}
\mathbf{I} &= (\mathbf{R} + \mathbf{\hat{R}}) \odot (\mathbf{L} + \mathbf{\hat{L}}) \\
	 &= \mathbf{R} \odot \mathbf{L} + \mathbf{R} \odot \mathbf{\hat{L}} + \mathbf{\hat{R}} \odot (\mathbf{L} + \mathbf{\hat{L}}).
\end{aligned}
\label{eq:retinex_noise}
\end{equation}
$\mathbf{\hat{R}} \in \mathbb{R}^{H\times W\times 3}$ and $\mathbf{\hat{L}} \in \mathbb{R}^{H\times W}$ are the perturbations. To light up $\mathbf{I}$, they multiply two sides of~\cref{eq:retinex_noise} by a light-up map $\mathbf{\bar{L}}$ such that $\mathbf{\bar{L}} \odot \mathbf{{L}} = \mathbf{1}$ as:
\begin{equation}
\mathbf{I} \odot \mathbf{\bar{L}} = \mathbf{R} + \mathbf{R} \odot (\mathbf{\hat{L}} \odot \mathbf{\bar{L}}) + (\mathbf{\hat{R}} \odot (\mathbf{L} + \mathbf{\hat{L}})) \odot \mathbf{\bar{L}},
\label{eq:retinex_light_up}
\end{equation}
They then simplify~\cref{eq:retinex_light_up} as
\vspace{-1mm}
\begin{equation}
\mathbf{I}_{{lu}}= \mathbf{I} \odot \mathbf{\bar{L}} = \mathbf{R} + \mathbf{C},
\label{eq:retinex_light_up_2}
\end{equation}
where $\mathbf{I}_{{lu}} \in \mathbb{R}^{H\times W\times 3}$ denotes the lit-up image and $\mathbf{C} \in \mathbb{R}^{H\times W\times 3}$ indicates the overall corruption term. Subsequently, they formulate their ORF as:
\begin{equation}
(\mathbf{I}_{{lu}}, \mathbf{F}_{{lu}}) = \mathcal{E}(\mathbf{I}, \mathbf{L}_{p}), ~~~~\mathbf{I}_{en} = \mathcal{R}(\mathbf{I}_{{lu}}, \mathbf{F}_{{lu}}), 
\label{eq:orf}
\end{equation}
where $\mathcal{E}$ denotes the illumination estimator and $\mathcal{R}$ represents the corruption restorer. $\mathcal{E}$ takes $\mathbf{I}$ and its  illumination prior map $\mathbf{L}_p \in \mathbb{R}^{H\times W}$ as inputs. $\mathbf{L}_{p}= \text{mean}_c(\mathbf{I})$  where $\text{mean}_c$ calculates the average pixel values across channels.
$\mathcal{E}$ outputs the lit-up image $\mathbf{I}_{lu}$ and light-up feature $\mathbf{F}_{lu} \in  \mathbb{R}^{H\times W\times C}$. Then $\mathbf{I}_{lu}$ and $\mathbf{F}_{lu}$ are  fed into $\mathcal{R}$ to restore the corruptions and produce the enhanced image $\mathbf{I}_{en} \in \mathbb{R}^{H\times W\times 3}$.

\noindent\textbf{Illumination Estimator.} The architecture of $\mathcal{E}$ is shown in~\cref{fig:pipeline} (a) (i).  $\mathcal{E}$ firstly uses a $conv$1$\times$1 (convolution with kernel size = 1) to fuse the concatenation of $\mathbf{I}$ and $\mathbf{L}_p$. The well-exposed regions can provide semantic contextual information for under-exposed regions. Thus, a depth-wise separable $conv$9$\times$9 is adopted to model the interactions of regions with different lighting conditions to generate the light-up feature  $\mathbf{F}_{lu}$. Then $\mathcal{E}$ uses a $conv$1$\times$1 to aggregate $\mathbf{F}_{lu}$ to produce the light-up map $\mathbf{\bar{L}} \in \mathbb{R}^{H\times W\times 3}$, which is used to light up $\mathbf{I}$ in~\cref{eq:retinex_light_up}. 

\noindent\textbf{Illumination-Guided Transformer.} As illustrated in~\cref{fig:pipeline} (a) (ii), IGT adopts a three-scale U-shaped architecture~\cite{cst,dauhst,bisci,rsn,sax-nerf}. The input of IGT is the lit-up image $\mathbf{I}_{lu}$. In the downsampling branch, $\mathbf{I}_{lu}$  undergoes a $conv$3$\times$3, an IGAB, a strided $conv$4$\times$4, two IGABs, and a strided $conv$4$\times$4 to generate hierarchical features $\mathbf{F}_{i} \in \mathbb{R}^{\frac{H}{2^i} \times \frac{W}{2^i}  \times 2^{i}C}$ where $i$ = 0, 1, 2. Then $\mathbf{F}_{2}$ passes through two IGABs. Subsequently, a symmetrical structure is designed as the upsampling branch. Skip connections are used to alleviate the information loss caused by the downsampling branch. The upsampling branch outputs a residual image $\mathbf{I}_{re} \in \mathbb{R}^{H\times W\times 3}$. Then the enhanced image $\mathbf{I}_{en}$ is derived by the sum of $\mathbf{I}_{lu}$ and $\mathbf{I}_{re}$, \emph{i.e.}, $\mathbf{I}_{en}$ = $\mathbf{I}_{lu}$ + $\mathbf{I}_{re}$.

\vspace{1mm}
\noindent\textbf{IG-MSA.} As illustrated in~\cref{fig:pipeline} (c), the light-up feature $\mathbf{F}_{lu} \in \mathbb{R}^{H\times W\times C}$ estimated by $\mathcal{E}$ is fed into each IG-MSA of IGT. Firstly, the input feature is reshaped into tokens $\mathbf{X} \in  \mathbb{R}^{HW\times C}$ and split into $k$ heads:
\begin{equation}
\mathbf{X} = [\mathbf{X}_1,~\mathbf{X}_2,~\cdots,~ \mathbf{X}_k], 
\label{split}
\end{equation}
where $\mathbf{X}_i \in \mathbb{R}^{HW\times d_k}, d_k = \frac{C}{k},$ and $i = 1, 2, \cdots, k$. Then $\mathbf{X}_i$ is linearly projected into $query$ $\mathbf{Q}_{i}$, \emph{key} $\mathbf{K}_i$, and \emph{value} $\mathbf{V}_i \in \mathbb{R}^{HW \times d_k}$ as: 
\begin{equation}
\mathbf{Q}_i = \mathbf{X}_i\mathbf{W}_{\mathbf{Q}_i}^{\text{T}},~~
\mathbf{K}_i = \mathbf{X}_i\mathbf{W}_{\mathbf{K}_i}^{\text{T}},~~
\mathbf{V}_i = \mathbf{X}_i\mathbf{W}_{\mathbf{V}_i}^{\text{T}},
\label{linear_proj}
\end{equation}
where $\mathbf{W}_{\mathbf{Q}_i}$, $\mathbf{W}_{\mathbf{K}_i}$, and $\mathbf{W}_{\mathbf{V}_i} \in \mathbb{R}^{d_k \times d_k}$ are learnable parameters and T denotes the matrix transpose. Subsequently, they use the light-up feature $\mathbf{F}_{lu}$ encoding illumination information and interactions of regions with different lighting conditions to direct the computation of self-attention. They reshape $\mathbf{F}_{lu}$ into $\mathbf{Y} \in \mathbb{R}^{HW\times C}$ and split it into $k$ heads:
\vspace{-0.5mm}
\begin{equation}
\mathbf{Y} = [\mathbf{Y}_1,~\mathbf{Y}_2,~\cdots,~ \mathbf{Y}_k], 
\label{split_Y}
\end{equation}
where $\mathbf{Y}_i \in \mathbb{R}^{HW\times d_k}, i = 1, 2, \cdots, k$. Then self-attention is formulated as:
\begin{equation}
\text{Attention}(\mathbf{Q}_i, \mathbf{K}_i, \mathbf{V}_i, \mathbf{Y}_i) = (\mathbf{Y}_i \odot \mathbf{V}_i)\text{softmax}(\frac{\mathbf{K}_{i}^{\text{T}}\mathbf{Q}_{i}}{\alpha_i}),
\label{attention}
\end{equation}
where $\alpha_i \in \mathbb{R}^1$ is learnable parameter. Subsequently, $k$ heads are concatenated to pass through an $fc$ layer and plus a learnable positional encoding $\mathbf{P} \in \mathbb{R}^{HW\times C}$ to produce the output tokens $\mathbf{X}_{out} \in \mathbb{R}^{HW\times C}$. Finally, they reshape $\mathbf{X}_{out}$ to derive output feature $\mathbf{F}_{out} \in \mathbb{R}^{H\times W\times C}$.

Besides Retinexformer~\cite{cai2023retinexformer}, the team also adopt their winning solution MST~\cite{mst,mst_pp} of NTIRE 2022 Spectral Recovery Challenge~\cite{arad2022ntire} as an auxiliary method for ensemble. The code is released at \url{https://github.com/caiyuanhao1998/MST} and \url{https://github.com/caiyuanhao1998/MST-plus-plus}.

\noindent\textbf{Implementation:} Retinexformer is implemented by PyTorch. The model is trained with the Adam~\cite{adam} optimizer ($\beta_1$ = 0.9 and $\beta_2$ = 0.999) for 2.5 $\times$ 10$^{5}$ iterations. The learning rate is initially set to 2$\times 10^{-4}$ and then steadily decreased to 1$\times 10^{-6}$ by the cosine annealing scheme~\cite{sgdr} during the training process. Patches at the size of 2000$\times$2000 are randomly cropped from the low-/normal-light image pairs as training samples. The batch size is 8. The training data is augmented with random rotation and flipping. The training objective is to minimize the mean absolute error (MAE) between the enhanced image and ground-truth image. The team use the mixed-precision training strategy (also the amp in Pytorch). 

In the testing phase, images with the size of 2000$\times$3000 are directly fed into the network. Images with the size of 4000$\times$6000 are firstly split into two 4000$\times$3000 images to undergo the model and then merged to obtain the enhanced images. Self-ensemble and multi-model ensemble strategies are used for final testing.

%% file: team03_DH_AISP/main.tex
\subsection{DH-AISP}
\noindent\textbf{Description:}
DWT-FFC \cite{zhou2023breaking} is employed as the backbone in our proposed method for low-light image enhancement. Dark light usually results in degraded image quality, low contrast, color shift, and structural distortion. We have observed that many deep learning-based models show superior performance in dark light enhancement, but they are generally not up to the challenge of lower light. There are two main factors contributing to this. First, due to the uneven distribution of image brightness, it is very challenging to recover structure and chromaticity features with highfdelity, especially in areas of extremely low brightness. Secondly, the existing small-scale data sets for dark light enhancement are not sufcient to support reliable learning of feature mapping based on convolutional neural network (CNN) models. To address these two challenges, we employ a novel two-branch network utilizing two-dimensional discrete Walsh transform (DWT), fast Fourier Convolution (FFC) residual blocks, and a pre-trained ConvNeXt model. Specifcally, in the DWT-FFC frequency branch, our model utilizes DWT to capture more high-frequency features. In addition, by utilizing the large receptive feld provided by FFC residuals, our model is able to efciently explore global context information and generate images with better perceptual quality. In the prior knowledge branch, we use the pre-trained ImageNet ConvNeXt instead of Res2Net. This allows our model to learn more complementary information and gain greater generalization ability. The feasibility and eﬀectiveness of the proposed method are proved by extensive experiments.

\noindent\textbf{Implementation:}
The training dataset is based solely on data provided by the competition organizers. The training images are generated by random clipping, and the training GPU is RTX 4080. A training session takes about 48 hours on a single GPU. The optimizer uses Adam. The initial learning rate is 0.0001, halved every 500 epochs.Training in python based on pytorch platform.

%% file: team04_NWPU_DiffLight/main.tex
\subsection{NWPU-DiffLight}
\begin{figure*}[!htp]
\includegraphics[width=1.0\linewidth]{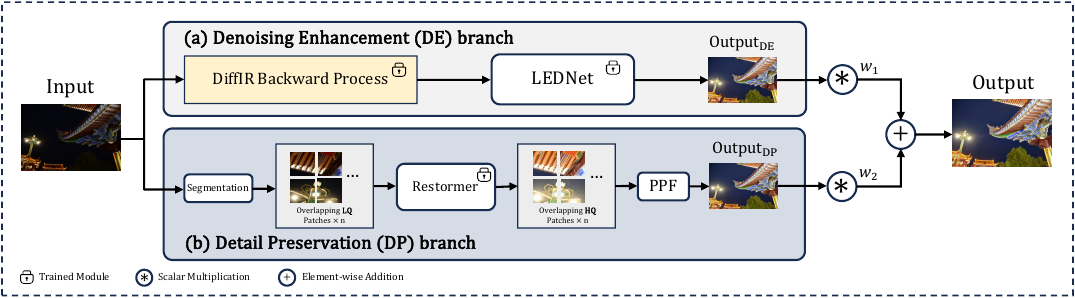}
\caption{DiffLight pipeline. DiffLight is composed of two branches: (a) Denoising Enhancement (DE) branch and (b) Detail Preservation (DP) branch.}
\label{fig:DiffLight}
\end{figure*}

\noindent\textbf{Description:~}As shown in~\cref{fig:DiffLight}, we proposed a dual-branch pipeline called DiffLight. Besides, we proposed a method, Progressive Patch Fusion (PPF) for high-resolution image restoration.
Given a low-light image as input, it will be fed into the two branches separately. In the DE branch (\cref{fig:DiffLight} (a) ), the low-quality image is processed sequentially by DiffIR \cite{diffir} and  LEDNet \cite{lednet} to get the image enhanced by DE branch, $Output_{DE}$:
\begin{equation}
\mathbf{Output}_{DE} = \text{LEDNet}(\text{DIffBackward}(\mathbf{Input})),
\end{equation}
where $\text{DiffBackward}(\cdot)$ represents the the Backward Process of DiffIR (which is for inference).
The enhanced image, $Output_{DE}$, has fairly low noise, minimal color deviation, as well as highly-increased brightness, but there is an excessive loss of details.

In the DP branch (\cref{fig:DiffLight} (b)), the newly proposed method for high-resolution image restoration, PPF, is used. To better recover the details from high-resolution images, the image is divided into $n$  small low-quality (LQ) overlapping patches in the Segmentation process, the patch size is adapted to the input size for training. Each patch is individually processed through the Restormer to obtain $n$ high-quality (HQ) overlapping patches. To get $Output_{DP}$, the image enhanced by DP-branch, PPF is applied on these enhanced patches: 
\begin{equation}
\mathbf{Output}_{DP} = \text{PPF}(\text{Restormer}(\text{Segmentation}(\mathbf{Input}))).
\end{equation}
The block artifacts produced by the traditional fusion method is removed by PPF, providing good visual quality and rich details.
Finally, $Output_{DE}$ and $Output_{DP}$ are multiplied by weights $w_1$ and $w_2$ respectively, and summed to form the final output image after weighted averaging:
\begin{equation}
\mathbf{Output} = w_1 \mathbf{Output}_{DE} + w_2 \mathbf{Output}_{DP}.
\end{equation}


 \noindent\textbf{Progressive Patch Fusion.} For high-resolution images, we utilize the Progressive Patch Fusion (PPF) method during testing. This approach incorporates progressive weight management at the to effectively mitigate edge and substantially improve visual fidelity. PPF is performed by these steps as follows:
\begin{enumerate}
    \item Image Segmentation: The input image $I$ is segmented into patches of size $p$. These patches slide across the image with a stride ($s$) of stride to determine their locations.
    \item Model Inference: The model is used to sequentially infer each patch, resulting in inferred patches.
    \item Weight Calculation: Four weight tensors, $weight_{1,2,3,4}$ are computed for blending overlapping regions. The weights for overlapping regions linearly range from 1 to 0. $weight_{1,2}$ are used for blending overlaps between patches, while $weight_{3,4}$ are used for merging overlaps between rows.
    \item Image Reconstruction: Initialize an empty restore image and $patch_{row}$ row image, which will be used to store the reconstructed image and the current row being processed.
    \item Fusion Process:
    \begin{itemize}
        \item For each $h_i$ (row index) and $w_i$ (column index), extract the corresponding patch from the patches tensor. If it is the first patch of the row ($w_i$ equals $0$), add it directly to the $patch_{row}$.
        \item For subsequent patches, use $weight_{1,2}$ to blend the overlapping regions between the two patches.
        \item Add the processed $patch_{row}$ to the restored image $I'$.
        \item If it is not the first row ($h_i \neq 0$), blending between rows using $weight_{3,4}$ is required.
    \end{itemize}
    \item Output: the aforementioned steps, restore tensor contains the the restored image $I'$.
\end{enumerate}

\noindent\textbf{Implementation: }
For the Denoising Enhancement (DE) branch, we train DiffIR \cite{diffir} and LEDNet \cite{lednet} separately. In training the diffusion model, total timesteps $T$ are set to $4$, and $\beta_t$ linearly increase from $\beta_1=0.9$ to $\beta_T=0.99$ . 
We train the two stage of DiffIR~\cite{diffir} only using $\mathcal{L}_{1}$ loss. We train $\text{DiffIR}_{S1}$ for 300K iterations with the initial learning rate $2\times10^{-4}$ gradually reduced  with the cosine annealing. 
And For $\text{DiffIR}_{S2}$, we train 300K iterations with initial learning rate $2\times10^{-4}$ and gamma $0.5$ with the MultiStepLR scheduler. 
For both training stage, we progressively increase patch size and decrease batch size. Specifically, during iterative training, the patch size and batch size pair are set to respectively train for $(92K)$, $(80K)$, $(38K)$, $(90K)$ iterations under the configurations of $(192,8)$, $(256,4)$, $(320,2)$, $(400,1)$.

LEDNet~\cite{lednet} model is trained on the inference results on Train set produced by DiffIR,   
and the Ground Truth remains unchanged. 
We train LEDNet using Adam~\cite{adam} optimizer with $\beta_{1}=0.9$, $\beta_{2}=0.99$ for a total of 300k iterations. The initial learning rate is set to $1\times10^{-4}$ and updated with cosine annealing strategy \cite{sgdr}. Similar to DiffIR, we still adopt a progressive training approach, the patch size and batch size pair are set to train for $(90K)$, $(70K)$, $(70K)$, $(70K)$ iterations respectively under the configurations of $(256,8)$, $(512,4)$, $(1024,1)$, $(1320,1)$.

We implement our the Detail-Preservation (DP) branch by PyTorch. The model is trained with the Adam \cite{adam} optimizer (\textit{$\beta_{1}$} = 0.9 and \textit{$\beta_{2}$} = 0.999) for at least 300 epochs by using a single NIVIDA 3090 GPU. The learning rate is initially set to $1 \times 10^{-4}$ and then steadily decreased to $1 \times 10^{-7}$ by the cosine annealing scheme \cite{sgdr} during the training process. We randomly crop the image to 256 $\times$ 256 for patch size and set batch size to $8$. When testing, we use our proposed Progressive Patch Fusion (PPF) method that addresses edge artifacts and yields favorable visual perceptual results with high resolution images.

%% file: team05_GiantPandaCV/main.tex
\subsection{GiantPandaCV}
\begin{figure*}[t]
 \centering
 \includegraphics[width=0.95\linewidth]{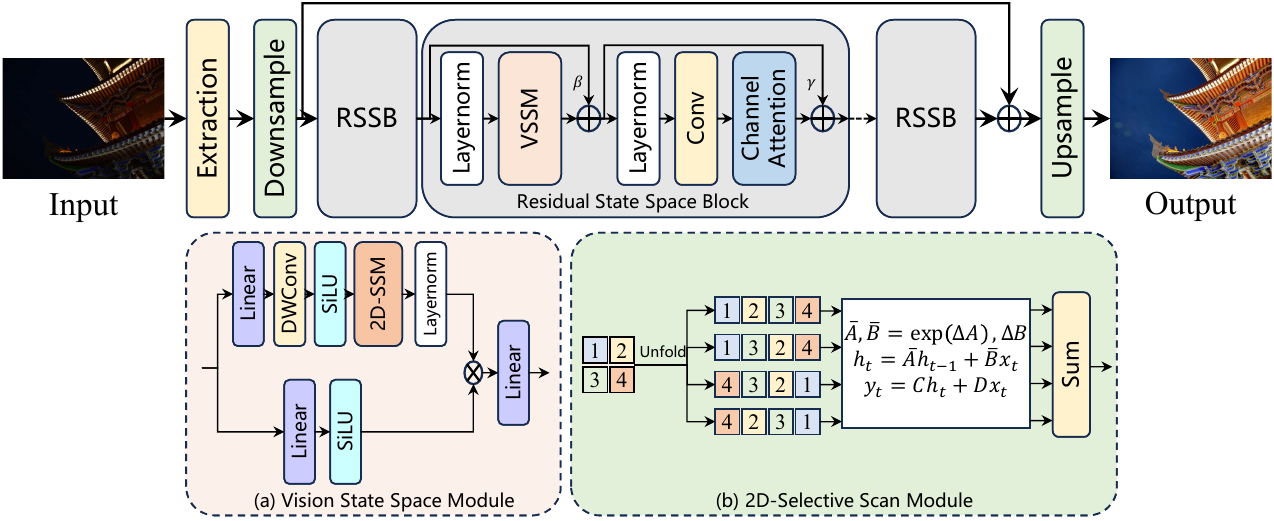}
 \caption{An overview of our UHDMamba for UHD Low Light Enhancement.}
 \label{fig: UHDMamba}
\end{figure*}
\noindent\textbf{Description:} 
Existing restoration backbones are usually limited due to the inherent local reductive bias or quadratic computational complexity in Ultra-High-Defnition (UHD) image restoration tasks. Recently, Selective Structured State Space Model, \eg, Mamba \cite{VMamba}, has shown great potential for long-range dependencies modeling with linear complexity. Therefore, we introduce the Selective Structured State Space Model \cite{Mamba} into the low light image enhancement task and propose an efficient model for UHD images, named UHDMamba. Specifically, our UHDMamba uses the Residual State Space Block as the core component, which employs convolution and channel attention to enhance the capabilities of the vanilla Mamba. To efficiently infer UHD images, we used PixelUnshuffle to downsample the UHD images and then input them into the network for enhancement. Finally, we used PixelShuffle upsampling to reconstruct the final image.

The main architecture of the UHDMamba is shown in~\cref{fig: UHDMamba}, which consists of three components: Shallow feature extraction, Deep feature enhancement, and Upsampling reconstruction. Given a low-quality (LQ) input image $I_{LQ} \in \mathcal{R}^{H\times W\times 3}$, we first employ a $3\times 3$ convolution layer from the shallow feature extraction to generate the shallow feature $F_{s} \in \mathcal{R}^{H\times W\times \frac{C}{S}}$, where H and W represent the height and width of the input image, C is the number of channels, and S is the downsample scale factor. Subsequently, the shallow feature $F_s$ undergoes downsample operation and the deep feature enhancement stage to acquire the deep feature $F_d^l \in \mathcal{R}^{\frac{H}{S} \times \frac{W}{S} \times SC}$ at the $l-$th layer, $l \in {1,2, ..., L}$. This stage is stacked by multiple Residual State-Space Blocks (RSSBs). Finally, we use the element-wise sum to obtain the input of the high-quality reconstruction stage $F_R = F_L + F_s$, which is used to Upsample reconstruct the high-quality (HQ) output image $I_{HQ}$.

\noindent\textbf{Implementation:}
The proposed method conducts experiments in PyTorch on two NVIDIA GeForce RTX 3090 GPUs. To optimize the network, the model employs the Adam \cite{adam} optimizer with a learning rate $2\times10^{-4}$. We randomly crop the full-resolution image to a resolution of $512\times 512$ as the input and perform 200k iterations of training with a batch size of 4. To augment the training data, random horizontal and vertical flips are applied to the input images. The number of RSSB and feature channels is set to 8 and 48, respectively. DownSampler and UpSampler are both composed of a sub-pixel convolutional layer.

In order to maximize the potential performance of our model, the method adopts the Test Time Augmentation strategy (TTA). During the test time, it used $90^\circ, 180^\circ, 270^\circ$ rotation, horizontal flip, and vertical flip to generate six augmented inputs $\{I_{n, i}^{input} = T_i(I_n^{input})\}$ from input left and right images $I_n^{{input}}$. With those augmented input images and the original input image, we generate corresponding clear images $\{I_{n,1}^{output}, \cdots, I_{n,6}^{output}\}$ using the networks. We then apply an inverse transform to those output images to get the original geometry $\tilde{I}_{n, i}^{output} = T_i^{-1}(I_{n, i}^{output})$. Finally, we average the transformed outputs altogether to make the TTA result as follows. $I_{n}^{output}=\frac{1}{6}\sum_{i=1}^{6}{\tilde{I}_{n, i}^{output}}$.

%% file: team06_LVGroup_HFUT/main.tex
\subsection{LVGroup\_HFUT}
\begin{figure}[!htp]
  \centering
   \includegraphics[width=1.0\linewidth]{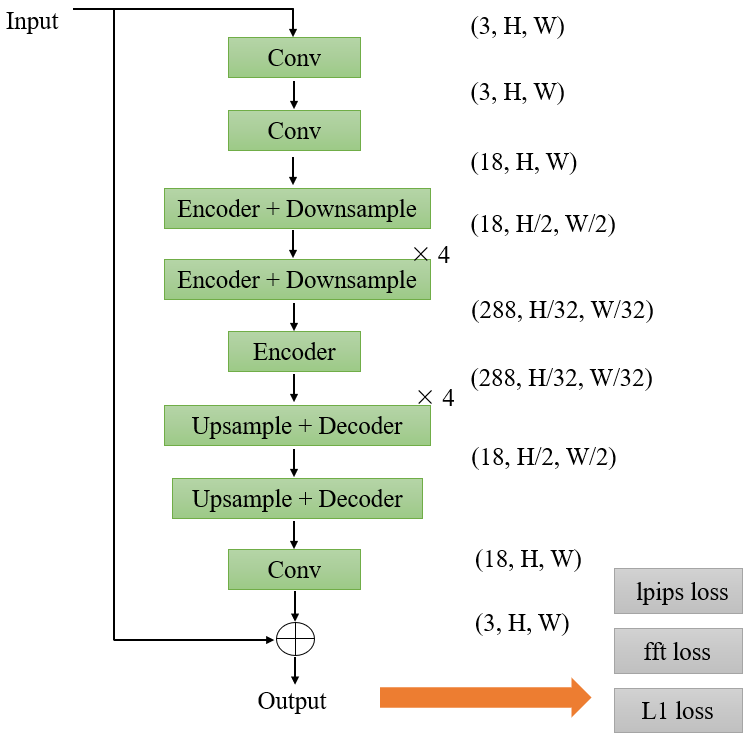}
   \caption{The network architecture of team LVGroup\_HFUT.}
   \label{fig:team_06_network}
\end{figure}
\noindent\textbf{Description:}~Low light enhancement refers to a process or set of techniques in computer vision aimed at improving the visual quality of images by adjusting their lighting conditions. This process can involve increasing the brightness and contrast, correcting underexposed or overexposed areas, and enhancing details in shadows and highlights. The goal is to produce images that more accurately reflect the scene as perceived by the human eye, even under less-than-ideal lighting conditions. However, complex models are difficult to reason about on the GPU, limited by the larger image resolution. Therefore, NAFNet \cite{chen2022simple} is chosed as baseline and further lighten NAFNet based on it to get a brand new structure. Specifically, the number of channels is kept constant at the early stage of the encoding phase and at the end of the decoding phase, while the up- and down-sampling operations are performed, as shown in~\cref{fig:team_06_network}.

More specifically, the input image is initially subjected to pixel alignment and convolution operations, during which the number of channels is expanded from 3 to 8. Subsequently, the image passes through an encoder and a downsampling module, resulting in an increase in channel count to 288, while its dimensionality is reduced to one-thirty-second of its original size. Following this, a decoder accompanied by an upsampling module is applied, which adjusts the channel count to 18 and restores the dimensions to their original size. The final output is then obtained through a convolutional layer. During the training process, the model is constrained using a combination of lpips loss, fft loss, and L1 loss to optimize performance and ensure fidelity to the input image.

\noindent\textbf{Implementation:}~The proposed architecture is based on PyTorch 2.2.1 and an NVIDIA 4090 with 24G memory. 2000 epochs are set for training with batch size 2, using AdamW with $\beta_1=0.9$ and $\beta_2=0.999$ for optimization.
The initial learning rate was set to 0.001, and cosine annealing was used for learning rate adjustment. The randomly crop the image to 768×768 is first performed and then horizontal flip with probability 0.5 is performed for data
augment. The input image is fed into the network, and it is constrained using three loss functions: lpips loss with weight 0.5, fft loss with weight 0.1, and $L_1$ loss with weight 1.

The testing process begins by combining a total of $10$ equally-spaced checkpoints obtained in training phase. Specifically, the input image with original resolution sequentially passes through these $10$ checkpoints, resulting in $10$ outputs. Finally, the model ensemble strategy is applied to process these $10$ outputs, yielding the ultimate output. Besides, since the test set contains images with too large resolution ($4000\times6000$), A100 (40G memory) is used for inference to avoid the `cuda out of memory' error.

%% file: team07_Try1try8/main.tex
\subsection{Try1try8}

\begin{figure}[!htp]
  \centering
   \includegraphics[width=1.0\linewidth]{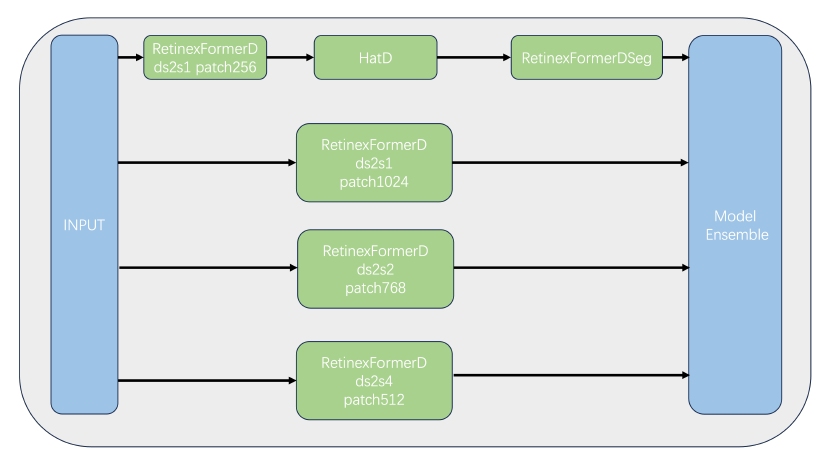}
   \caption{The network architecture of team Try1try8.}
   \label{fig:team_07_network}
\end{figure}

\noindent\textbf{Description:} Due to the large resolution of the inference images, the authors added a downsampling operation to
the RetinexFormer, renaming it to RetinexFormerD. The authors employed a RetinexFormerD model architecture with varying training patch sizes, and cascaded it with a denoising model (HatD)
and a fine-tuning model(RetinexFormerDseg). Finally, the authors performed a model ensemble of all the model results. The model is shown in~\cref{fig:team_07_network}.

\noindent\textbf{Implementation:~}\textit{path1:~}Initially, the authors applied downsampling to the RetinexFormer to create RetinexFormerD (due to the large size of the inference images), followed by the use of the HATD model (which incorporates
the HAT model with downsampling) to denoise the inference results of RetinexFormerD. Ultimately, RetinexFormerSeg (which integrates image segmentation results into the model) was utilized to fine-tune the final results. \textit{path2:~}The authors trained RetinexFormerD with various patch sizes and stages, resulting in three Retinex-FormerD models: RetinexFormerD ds2s1 with a patch size of 1024, RetinexFormerD ds2s2
with a patch size of 768, and RetinexFormerD ds2s4 with a patch size of 512. Finally, the authors averaged the results obtained from path1 and path2. The model is trained with Adam optimizer, 2e-4 as learning rate, and on V100

%% file: team08_Pixel_warrior/main.tex
\subsection{Pixel\_warrior}
\begin{figure}[!htp]
  \centering
   \includegraphics[width=\linewidth]{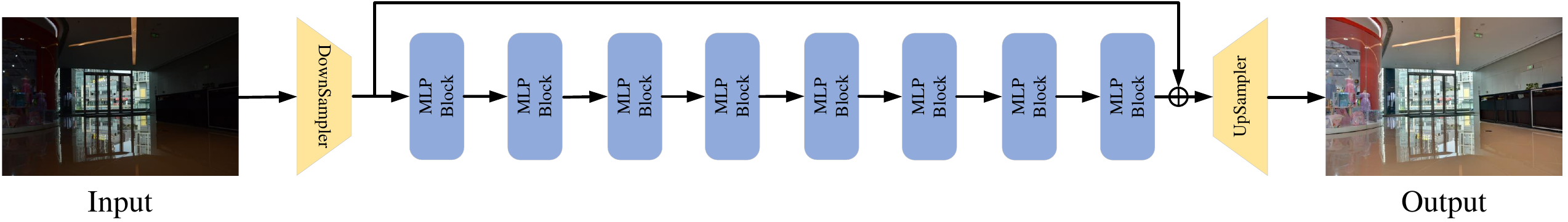}
   \caption{The network architecture of team Pixel$\_$warrior.}
   \label{fig:team08_network}
\end{figure}
\begin{figure}[t]
  \centering
   \includegraphics[width=\linewidth]{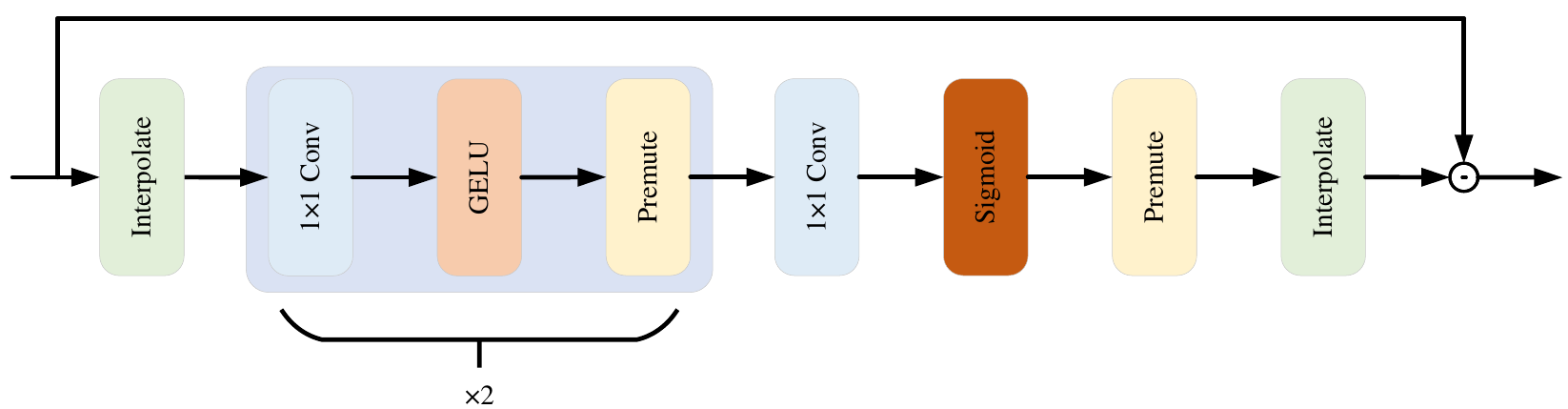}
   \caption{The network architecture of the MLP block.}
   \label{fig:team08_network_GFEM}
\end{figure}

\noindent\textbf{Description:}
The team proposes an efficient vision MLP-based architecture for low-light enhancement (see~\cref{fig:team08_network}). Our MLP block contains dimension transformation operations (see Fig.~\ref{fig:team08_network_GFEM}). Specifically, we first normalize the input features, and then perform multi-stage dimension transformations to rotate the spatial perspective of tensors across 3 dimensions of $H$, $W$ and $C$. Here, the 3D feature map undergoes recursive encoding from $(C, H, W)$ to $(H, W, C)$ and then to $(W, C, H)$, enabling the capture of global spatial information through multi-view dimensions. Finally, we adjust the feature map to the original resolution, and interact with the input features to activate useful features.

\noindent\textbf{Implementation:}
We conduct network training on four NVIDIA Tesla V100 GPUs with 32GB memory. In total, we perform 500 epochs of training. During the training, we adopt the Adam optimizer with a learning rate of $2 \times 10^{-4}$. The patch size is set to be $2000 \times 2000$ pixels and the batch size is set to be 4. To augment the training data, we apply random horizontal and vertical flips.

%% file: team09_HuiT/main.tex
\subsection{HuiT}
\begin{figure}[!htp]
\centerline{\includegraphics[width=1.0\linewidth]{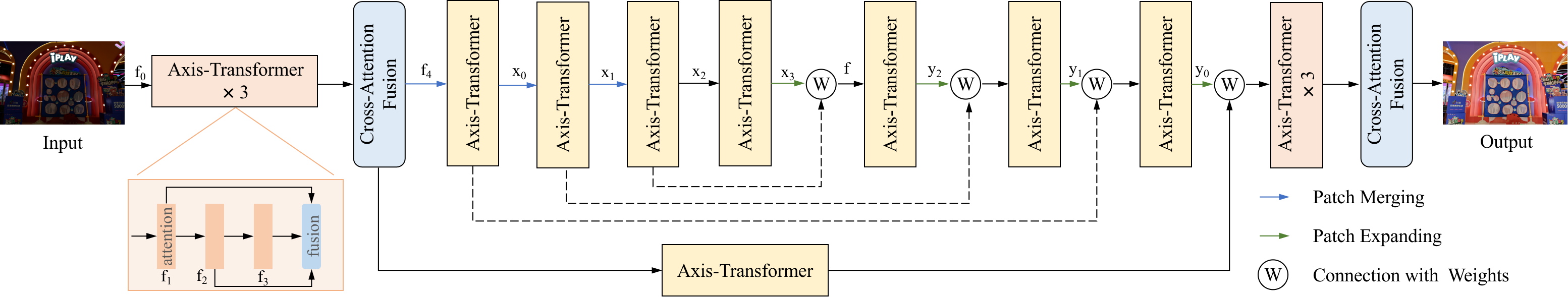}}
\caption{Illustration of the propsed LLformer.(Re-produced from LLformer~\cite{wang2023ultra})}
\label{ATL}
\end{figure}
\begin{figure}[t]
\centerline{\includegraphics[width=1.0\linewidth]{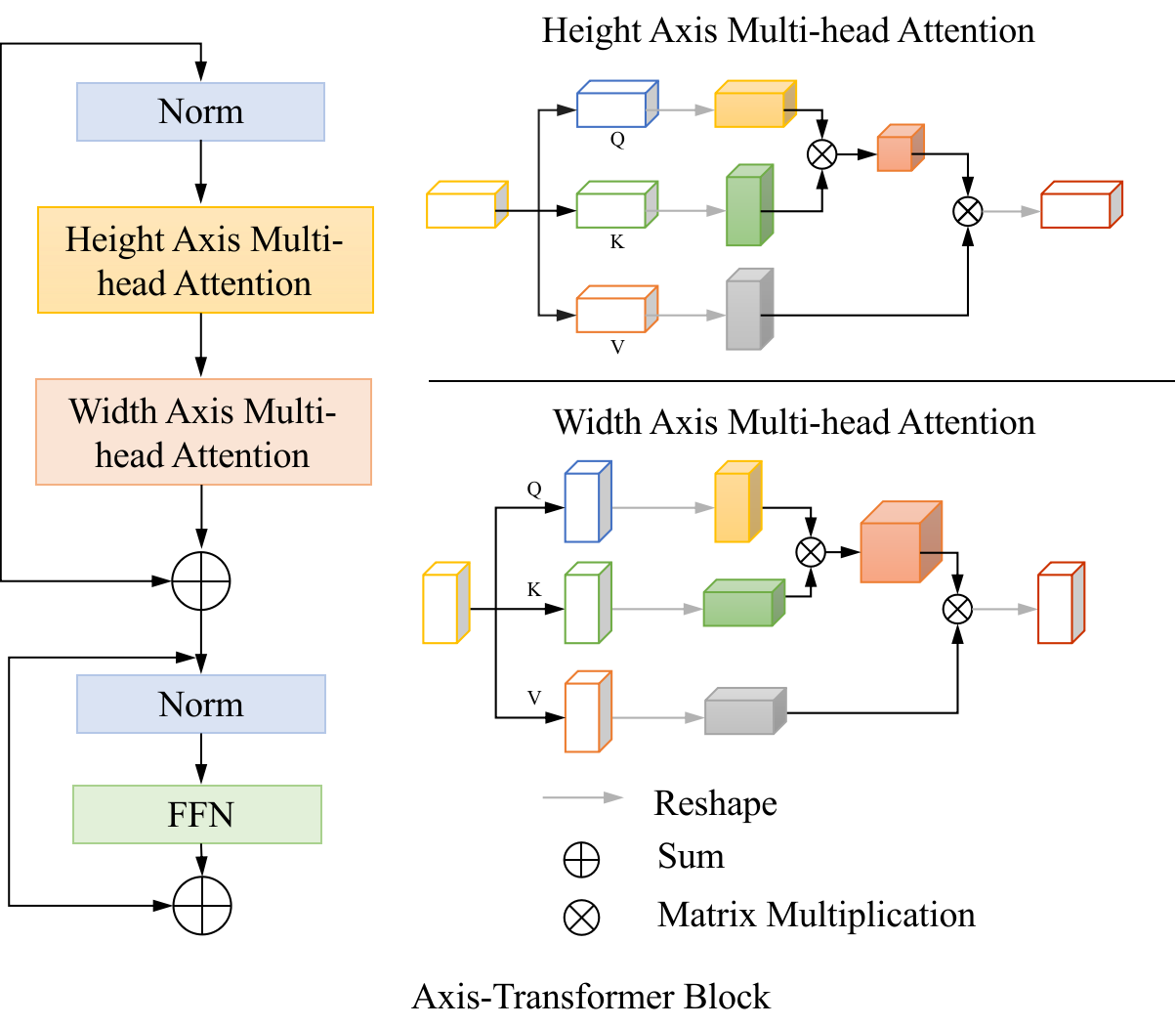}}
\caption{Illustration of the proposed Axis-Transformer Block.}
\label{ATB}
\end{figure}
\begin{figure}[t]
\centerline{\includegraphics[width=1.0\linewidth]{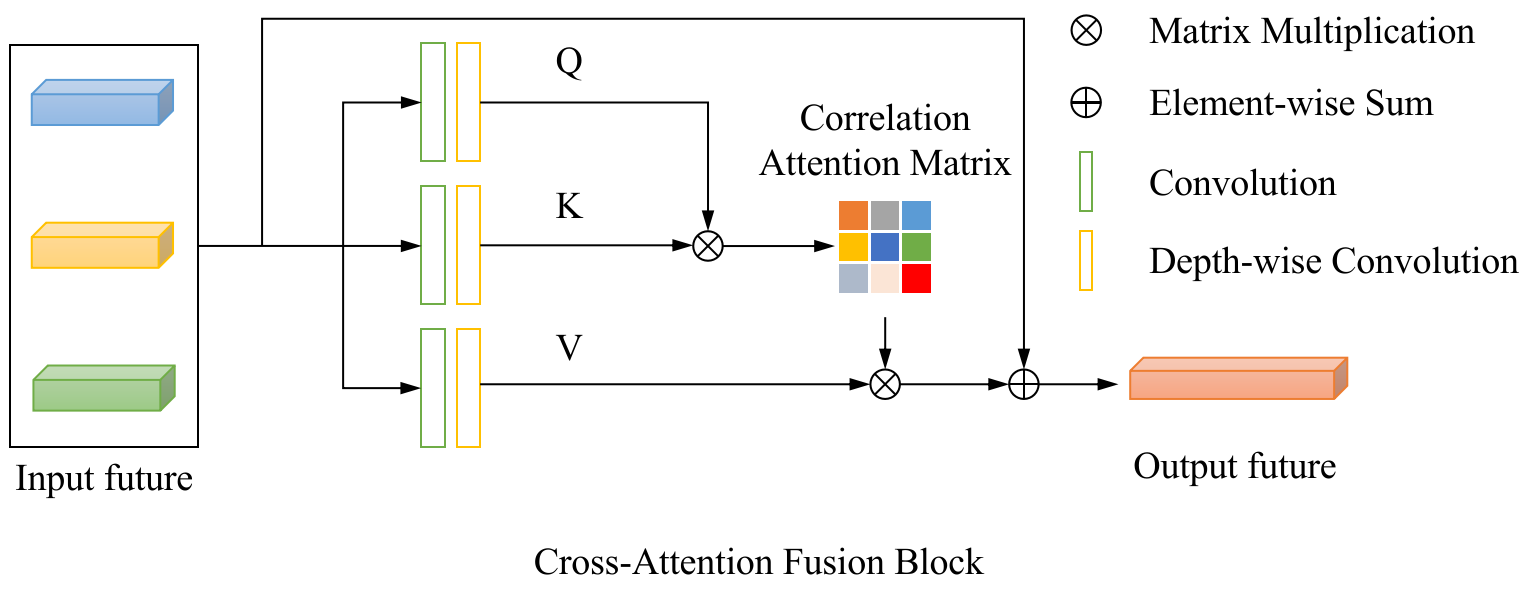}}
\caption{Illustration of the proposed Cross-Attention Fusion Block.}
\label{CAFB}
\end{figure}

\medskip\noindent\textbf{Description:} The team introduced LLformer~\cite{wang2023ultra} to address this issue. As shown in \cref{ATL}, given a low-light image, LLformer first extracts shallow features $f_0$, which are then fed into three sequential Transformer blocks to extract deep features. Specifically, the intermediate features output from the Transformer blocks are denoted as $f_1$, $f_2$, $f_3$. These features are aggregated and transformed into enhanced image features $f_4$ using the proposed cross-layer attention fusion block. Subsequently, deep feature extraction is performed on $f_4$ using four stages in the encoder. Specifically, each stage includes Patch Merging and Axis-Transformer blocks. Then, the low-resolution latent features are gradually restored to a high-resolution representation using the decoder, which consists of three stages and takes $x_3$ as input. Each stage consists of Patch Expanding and Axis-Transformer blocks. To reduce information loss at the encoding end and achieve better feature restoration at the decoding end, weighted skip connections with $1 \times 1$ convolutions are used for feature fusion in both the encoding and decoding ends. Thirdly, after decoding, the deep features f are sequentially processed through three Axis-Transformer blocks and a cross-layer attention fusion block to generate enhanced features for image reconstruction. Finally, the enhanced image is produced. The Axis-Transformer block, as shown in ~\cref{ATB}, performs self-attention mechanism along the height and width axes of the features in the cross-channel dimension to capture non-local self-similarity and long-range correlations with lower computational complexity. The Cross-Attention Fusion block, as shown in~\cref{CAFB}, learns attention weights between different layer features and adaptively fuses features with the learned weights to improve feature representation.

\medskip\noindent\textbf{Implementation:}~The code is implemented based on the Pytorch framework, We trained the model in 300 epochs on two NVIDIA 4090 GPU with batchsize of $24$. The model are optimized by the Adam with $\beta_1$ = $0.9$ and $\beta_2$ = $0.999$ with weight decay 1e-8 by default. The initial learning rate was set to 2$e$-4, and true cosine annealing was chosen as the learning scheme. The loss function is defined as:
\begin{equation}
    \pounds_{total} = \pounds_{pix} + \lambda \pounds_{ssim},
\end{equation}
where $\pounds_{pix}$ represents the L1 Loss, and $\pounds_{ssim}$ signifies the SSIM Loss. The parameter $\lambda$ denotes the loss weight used to balance the influence between the two terms. In our experiments $\lambda$ is set to $1.0$.

To save computational resources, high-definition images need to be chunked, however, the size of the patches has a great impact on the results. Experiments have shown that the best results will be achieved by segmenting the original image of size $2000 \times 2996$ into multiple patches of size $500 \times 748$, and by segmenting the original image of size $4000 \times 6000$ into patches of size $800 \times 1200$. We randomly cropped $128 \times 128$ smaller patches from the processed images during the network training process. Following the data augmentation techniques outlined in the LLFormer~\cite{wang2023ultra} methodology, we applied random horizontal and vertical flips to these patches.

%% file: team10_X_LIME/main.tex
\subsection{X-LIME}
\noindent\textbf{Description:}~Our method consists of two modules, which are the curve estimation network for lightness adaption and the denoising network for denoising. Specifically, we adopt Zero-DCE~\cite{guo2020zero} as a lightness adaption network and NAFNet \cite{DBLP:conf/eccv/ChenCZS22} as a denoising network. For better lightness adaption performance, we adjust the relevant loss function. Specifically, we add MSELoss to the original Zero-DCE~\cite{DBLP:conf/cvpr/GuoLGLHKC20} loss and train for 10 epochs on the provided dataset. In the training phase, we use Zero-DCE pre-trained weights provided in the official Github repository and fine-tune the provided dataset for the first stage. For the second stage, we used NAFNet~\cite{DBLP:conf/eccv/ChenCZS22} pre-trained on the SIDD dataset and fine-tuned the output of the lightness module. In the testing phase, we first use the lightness adaption module to adapt the lightness, thus increasing the contrast in the image. Then we feed the output of the lightness module into the denoising module to further mitigate the noise.

\noindent\textbf{Implementation:}~Our proposed method is implemented in Python with the help of the PyTorch framework. The models used in the two stages are optimized with Adam. In the first stage, the learning rate is set to 1e-4 and in the second stage, the learning rate is set to 5e-5, respectively. In the first stage, we adapt the RTX4090 GPU while the A100 80G is used in the second stage. The convergence speed of the first stage is rather fast, so we only train 10 epochs on the training set with the full image as input, resulting in a 30-minute training. For the second stage, we train the NAFNet~\cite{DBLP:conf/eccv/ChenCZS22} with a progressively enlarged patch strategy. The first 30000 iterations are trained with $512 \times 512$ patches, then we enlarge the training patch to $1024\times1024$ for 25000 iterations. Finally, we use $1920\times1920$ patches for 10000 iterations. During testing, we use TLC~\cite{DBLP:conf/eccv/ChuCCL22} to further boost the performance of our patch-based method. The base size of TLC is set to (1920, 2980).

%% file: team11_Image_Lab/main.tex
\subsection{Image Lab}
\begin{figure}[!htp]
  \centering
     \includegraphics[width=1.0\linewidth]{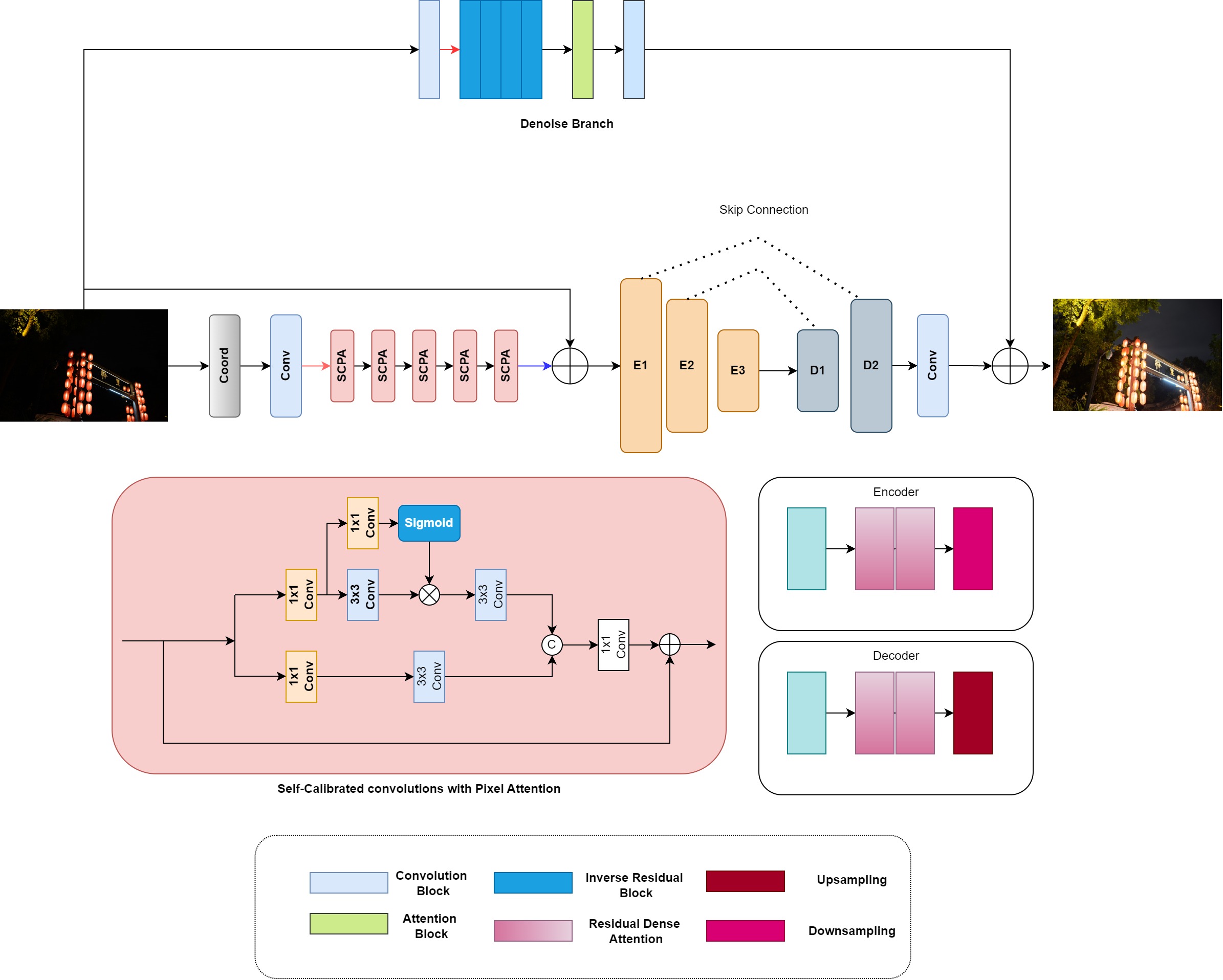}
  \caption{Architecture Diagram of ImageLab Team's Demosaicing Model}
  \label{fig:team11_architecture}
\end{figure}
\noindent\textbf{Description:}
We propose a novel image denoising and enhancement architecture, leveraging the power of Coordinate Convolution (CoordConv)~\cite{Liu2018} layers and Self-Calibrated Convolution with pixel attention (SCPA) blocks~\cite{Zhang2022_image,Kansal2023}, as shown in~\cref{fig:team11_architecture}. The architecture is designed to suppress noise while effectively preserving important image details.

The input image is first passed through a CoordConv layer to add channels containing hard-coded coordinates, enriching the representation with spatial information. This augmented representation is then downsampled and fed into five consecutive SCPA layers with a Pixel Attention Block. SCPA layers enhance the model's capture of intricate spatial patterns and features. 
The output from the SCPA layers is upsampled and added back to the input image. This combined representation is passed through a Modified U-Net architecture with Residual Dense Channel Attention (RDCA) blocks~\cite{Uma2020}.~The U-Net consists of three encoder blocks and two decoder blocks. Each encoder block contains two RDCA blocks followed by downsampling, while each decoder block contains two RDCA blocks followed by upsampling. This design facilitates the extraction and refinement of features at multiple scales. Simultaneously, the input image is fed into a sophisticated denoising block~\cite{Vasluianu2023} to produce a three-channel denoised image. The denoising block comprises four inverse convolutional layers followed by an attention mechanism. This mechanism effectively suppresses noise while preserving important image details, enhancing the overall quality of the denoised image. The outputs of the RDCA-UNet and the denoising block are added to the input image, resulting in a final enhanced image combining the denoising block's denoised features with the refined features from the RDCA-UNet.

\noindent\textbf{Implementation:}
The proposed network was trained using the NVIDIA Tesla P100 with 16GB RAM and the TensorFlow Keras platform. 400x400x3 patches were randomly extracted from the images. The training dataset consisted of 4281 patches, and 755 patches were used for validation. Augmentation techniques were applied during training. The model was optimized using the Adam optimizer with a learning rate schedule that decreased from 0.001 to 0.00001 over 250 epochs. The model comprises 0.437556 million parameters. The training objective included a combination of $\mathcal{L}_1$, $ \mathcal{L}_{SSIM}$, and $\mathcal{L}_{Grad}$ loss, \ie,
\begin{equation}
\label{eqn:image_lab}
    \mathcal{L} = 0.1*\mathcal{L}_{SSIM}+\mathcal{L}_1+\mathcal{L}_{Grad}.
\end{equation}

%% file: team12_dgzzqteam/main.tex
\subsection{dgzzqteam}
\begin{figure*}[!htp]
    \centering
    \includegraphics[width=1.0\linewidth]{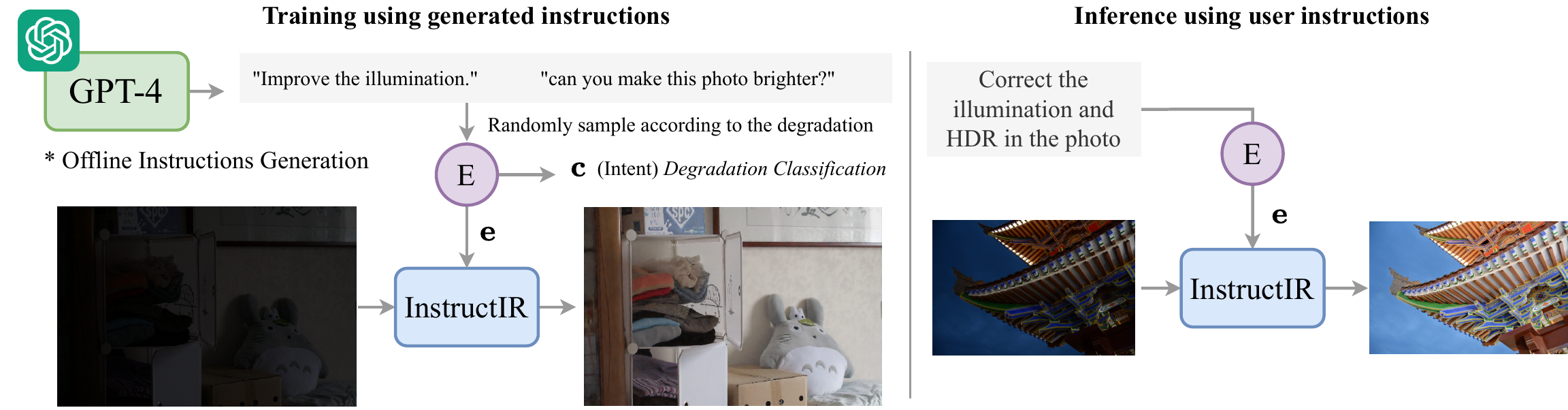}
    \caption{\textbf{InstructIR}~\cite{conde2024high} takes as input an image and a human-written instruction for how to improve that image. The multi-task model performs text-guided low-light image enhancement.}
    \label{fig:cidautai@instructir}
\end{figure*}
\noindent\textbf{Description:} 
The author claimed that the solution is based on the published OGF, while not providing the specific citation. The authors changed the loss function and fine-tuned the model so that it could perform better in the competition.

\noindent\textbf{Implementation:} The authors implement the OGF by PyTorch. The model is trained with the Adam optimizers ($\beta_1$ = 0.9 and $\beta_2$ = 0.999) for 2.5 $\times$ 105 iterations. The learning rate is initially set to 2$e$-4 and then steadily decreased to 1$e$-6 by the cosine annealing scheme during the training process.
Patches at the size of 128 $\times$ 128 are randomly cropped from the low-/normallight
image pairs as training samples. The batch size is 8. The training objective
is to minimize the mean absolute error (MAE) between the lit-up image and
the enhanced image. The training and validation sets are split in proportion to 209:21.

%% file: team13_Cidaut_AI/main.tex
\subsection{Cidaut AI}
\noindent\textbf{Description:~}~We use \textbf{InstructIR}~\cite{conde2024high} for real-world low-light enhancement. InstructIR~\cite{conde2024high} takes as input an image and a human-written instruction for how to improve that image. The neural model performs all-in-one image restoration. InstructIR~\cite{conde2024high} achieves state-of-the-art results on several restoration tasks including image denoising, deraining, deblurring, dehazing, and (low-light) image enhancement. The approach is illustrated in \cref{fig:cidautai@instructir}.

The model achieves $23.07$ dB, $0.8075$ SSIM and $0.156$ LPIPS in the NTIRE 2024 Low-light Enhancement Challenge, \emph{representing a baseline solution for multi-task restoration using text-guidance.}

\medskip\noindent\textbf{Implementation:}~The model can process full-resolution images in a regular GPU without tiling strategies. We fine-tune the model using the challenge dataset besides LOL~\cite{wei2018deep}. We use the instructions \emph{``correct the low illumination in this image"} for all the test images.

\vspace{2mm}

\noindent\textit{Efficient Baseline Methods:} The team also studies efficient methods that can process full-resolution images in real-time at several FPS: RetinexNet~\cite{wei2018deep}, SCI~\cite{ma2022toward} and Zero-DCE~\cite{guo2020zero}. All these methods can process $4000\times6000$ directly on regular GPUs such as Nvidia 3090Ti. However, these methods did not improve over InstructIR~\cite{conde2024high}.

\vspace{2mm}

\noindent\textit{Datasets:} To enhance our results and generalize better, we included other training datasets: LOL-v2 (real and synthetic) and the MIT5K dataset. We also apply random crop, flip and rotation augmentations.

%% file: team14_OptDev/main.tex
\subsection{OptDev}
\begin{figure}
    \centering
    \includegraphics[width=0.95\linewidth]{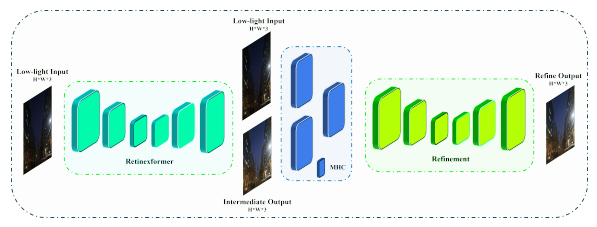}
	\caption{Overall framework the Team OptDev's DFormer}
	\label{fig:optdev-model}
\end{figure}
\noindent\textbf{Description:} The authors developed a novel transformer-based deep network DoubleFormer (DFormer) to learn
low-light to highlight mapping, as shown in \cref{fig:optdev-model}. The architecture comprises two separate
encoder-decoder blocks (EDB) and a multi-head correlation block (MHCB) to produce plausible images. The authors leverage illumination mapping from the well-known Retinex theory to accelerate
our reconstruction performance. Also, the authors incorporated the state-of-the-art low-light enhancement method, Retinexformer in the overall architecture’s first half. Regrettably, Retinexformer failed
to generate plausible images in many tricky scenes while illustrating visible noise and color desaturation in complex regions. To address this limitation, the authors proposed to utilize an MHCB, followed by
another EDB in our architecture. We leverage the correlated features with intermediate output in the second EDB to perceive better denoising results. In addition to that, the authors utilized a perceptual loss, including luminance-chrominance guidance, to address the color inconsistency.

\noindent\textbf{Implementation:} The proposed solution is implemented with the PyTorch framework. The networks were optimized with a Adam optimizer, where the hyperparameters were tuned as $\beta_1$ = 0.9, $\beta_2$ = 0.99, and
learning rate = 5e-4. We trained our model with non-overlapping image patches with a constant batch size of 12, which takes around 36 hours to complete. We conducted our experiments on an NVIDIA A6000 graphical processing unit (GPU) machine.

%% file: team15_ataza/main.tex
\subsection{ataza}
\begin{figure*}
	\centering
\includegraphics[width=1.0\linewidth]{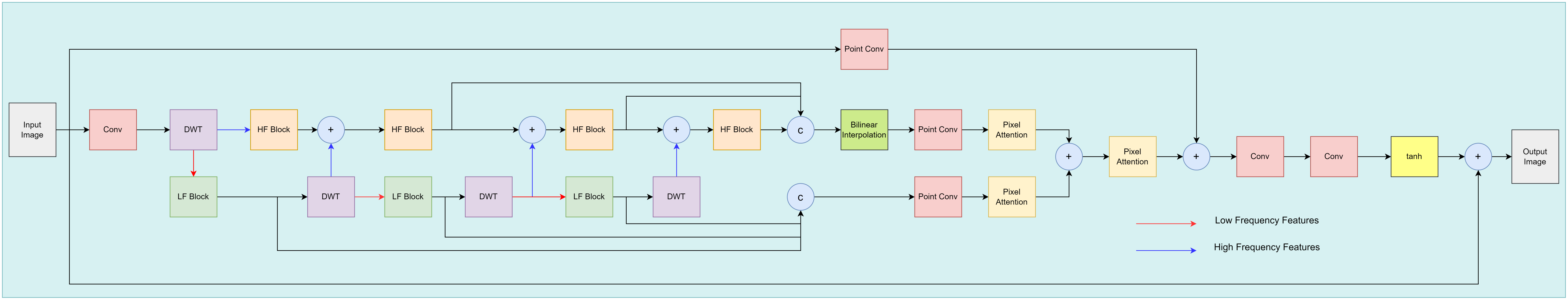}
	\caption{Overall framework our Team ataza's Frequency Guided Network}
	\label{fig:ataza-model}
\end{figure*}
\noindent\textbf{Description:~}Team ataza proposes a low parameter efficient network, as shown in~\cref{fig:ataza-model}. 
Based on the observation that error in low light images mainly exist in the low-frequency region they decompose the image into high-frequency and low-frequency subbands using Discrete Wavelet Transform (DWT) whose utility for image processing has been well established~\cite{parkMotionEstimationUsing2000}. Specifically, they choose HAAR wavelet~\cite{mallatTheoryMultiresolutionSignal1987} due to it's computational efficiency and excellent ability to detect edges~\cite{mashfordEdgeDetectionPipe2014}. The model consists of two branches one for high-frequency and the other for low-frequency features. This approach allows the model to focus on recovering low-frequency information while preserving high-frequency information that otherwise would have been lost in the deeper layers.

In the low-frequency branch, three Low-Frequency Blocks (LF Blocks) handle and extract low-frequency features. Subsequently, after each LF Block, a DWT block re-divides the output, channeling the high-frequency features to the high-frequency branch. Within the high-frequency branch, the existing high-frequency features are fused with those received via the DWT block from the low-frequency branch, before undergoing processing by the High-Frequency Block (HF Block). A total of four HF Blocks are employed in this process. This approach enables the model to preserve high-frequency features while concentrating on the recovery of low-frequency information. Given that the low-frequency subbands of the image have more errors, our approach involves making the LF Block deeper and more complex compared to the HF Block.

The outputs of the three LF Blocks are concatenated, while the outputs of the last three HF Blocks are also concatenated. Both sets of features are then fused and the final image generated. The model is trained using two loss functions: the $L_1$ loss and a frequency-domain-based loss $Loss_{FFT}$ derived from Fast Fourier Transform (FFT). $Loss_{FFT}$ is defined by \cref{eq:ataza_eq1}, and the overall loss is determined by \cref{eq:ataza_eq2}. Incorporating a frequency-based loss enhances model stability since the network is extracting features in the frequency domain, \ie,
\begin{equation}
	Loss_{FFT} = \frac{1}{n} \sum_{i=1}^{n} | |FFT(I_g)| - |FFT(I_t )||,
	\label{eq:ataza_eq1}
\end{equation}
\begin{equation}
	Loss = Loss_{FFT} + 20*Loss_{L_1}.
	\label{eq:ataza_eq2}
\end{equation}

This approach enables the solution to possess only 387,414 parameters while still achieving competitive performance. It can process a 1024x1024 image in just 0.14s and a 512x512 image in 0.04s. The rapid speed, coupled with low complexity, renders it suitable for various real-world applications where a balance of performance and speed is crucial.

\noindent\textbf{Implementation:}
The solution, developed in Python with the PyTorch framework, was trained on an Intel i7-8700 CPU @ 3.20GHz, 16GB RAM, and NVIDIA GeForce RTX 3070 graphics card. Data provided by the competition was exclusively used for training the model. The training dataset comprised 230 pairs of images, each sized 2992x2000 pixels. Out of these, 220 images were allocated for training, with the remaining 10 reserved for validation. To augment the dataset, non-overlapping patches of size 256x256 pixels were extracted from each image and used to train the model. Training lasted 70 epochs, employing the Cosine Annealing strategy with a cycle length of 10 epochs and an initial learning rate set at 0.0005. To train the model Adam optimizer, in conjunction with PyTorch's mixed precision feature was used and batch size was set to 8.

%% file: team16_KLETech_CEVI_LowlightHypnotise/main.tex
\subsection{KLETech-CEVI\_LowlightHypnotise}
\begin{figure}[!htp]
    \includegraphics[width=1\linewidth]{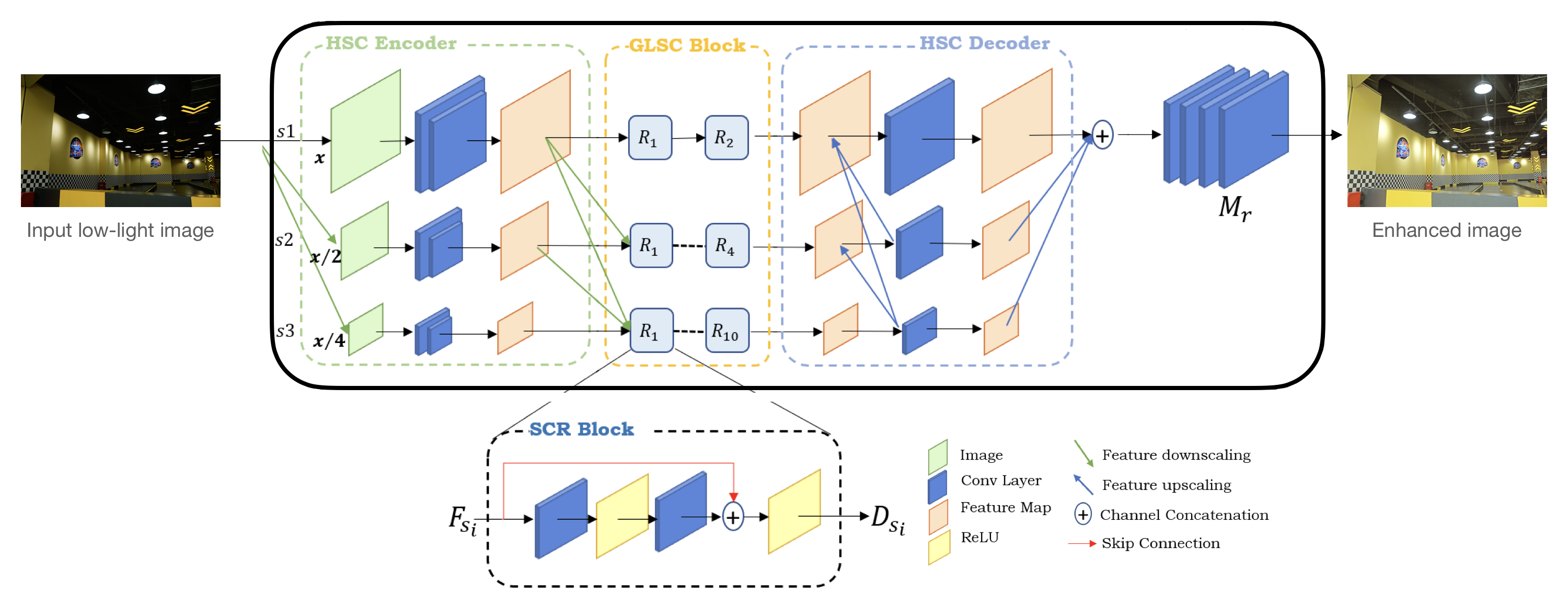}
    \caption{Overview of the proposed Multi-scale Feature FusionNet (MFNN). The encoder extracts features in three distinct scales, with information passed across hierarchies \textcolor{green}{(green dashed box)}. Fine-grained global-local spatial and contextual information is learnt through the GLSC Block \textcolor{orange}{(orange dashed box)}. At decoder, information exchange occurs in reverse hierarchies \textcolor{blue}{(blue dashed box)}.}
    \label{fig:mfnn-arch}
\end{figure}
\noindent\textbf{Description:}~The proposed MFNN framework includes three main modules: the hierarchical spatio-contextual (HSC) feature encoder, Global-Local Spatio-Contextual (GLSC) block, and hierarchical spatio-contextual (HSC) decoder, as shown in \cref{fig:mfnn-arch}. Typically, low-light image enhancement networks employ feature scaling for varying the sizes of the receptive fields. The varying receptive fields facilitate learning of local-to-global variances in the features. With this motivation, we learn contextual information from multi-scale features while preserving high resolution spatial details. We achieve this via a hierarchical style encoder-decoder network with residual blocks as the backbone for learning.
Given a input image x, the proposed multi-scale hierarchical encoder extracts shallow features in three distinct scales and is given as: 
\begin{equation}
    F_{si} = M_{Es}(x),
\end{equation}
where $F_{si}$ are the shallow features extracted at the $i^{th}$ scale from the sampled space of input image $x$ and, $M_{Es}$ represents the hierarchical encoder at scale $s$.
To learn the global-to-local representations from these shallow-level features, we propose a Global-Local Spatio-Contextual (GLSC) block, with residual blocks as the backbone. The learnt deep features are represented as:
\begin{equation}
    D_{si} = \text{GLSC}_{si}(F_{si}),  
\end{equation}
where $D_{si}$ is the deep feature at the $i^{th}$ scale, $F_{si}$ are the shallow features extracted at the $i^{th}$ scale and, $GLSC_{si}$ represents residual blocks at respective scales.
We decode the deep features obtained at various scales, with the proposed hierarchical decoder and is given by:
\begin{equation}
    d_{si} = M_{dsi}(D_{si}),
\end{equation}
where \(d_{si}\) is the decoded feature at the \(i^{th}\) scale, and \(M_{dsi}\) represents the hierarchical decoder.
The decoded features and upscaled features at each scale are passed to the reconstruction layers \(M_r\) to obtain the enhanced image \({\hat{y}}\). The upscaled features from each scale are stacked and represented as:
\begin{equation}
    P = d_{s1} + d_{s2} + d_{s3}, 
\end{equation}
where \(d_{s1}\), \(d_{s2}\), and \(d_{s3}\) are decoded features at three distinct scales, \(P\) represents the final set of features passed to reconstruction layers to obtain the enhanced image \({\hat{y}}\):
\begin{equation}
    \hat{y} = M_r(P),
\end{equation}
 where \(\hat{y}\) is the enhanced image obtained from reconstruction layers \(M_r\).
We optimize the learning of MFNN with the proposed ${\mathcal{L}_{MFNN}}$ and is given as:
\begin{equation}
    \mathcal{L}_{MFNN} = \alpha * L_{1} + \beta *\mathcal{L}_{VGG} + \gamma * \mathcal{L}_{MSSSIM},
    \label{hnnloss}
\end{equation}
where $\alpha$, $\beta$, and $\gamma$ are the weights. We experimentally set the weights to $\alpha=0.5$, $\beta=0.7$, and $\gamma=0.5$.
${\mathcal{L}_{MFNN}}$ is a weighted combination of three distinct losses inspired from \cite{desai2023lightnet,hegde2022ae, Desai_2023_WACV, Desai_2022_CVPR}. $L_1$ loss to minimize error at pixel level, perceptual loss \cite{vggloss} to efficiently restore contextual information between the ground-truth image and the output image, multiscale structural dissimilarity loss to restore structural details. The aim here is to minimize the weighted combinational loss  ${\mathcal{L}_{MFNN}}$ given as:
\begin{equation}
    L(\theta) = \frac{1}{N} \sum_{i=1}^{N} \parallel MFNN(x_i - y_i) \parallel_{\mathcal{L}_{MFNN}},
\end{equation}
where \(\theta\) denotes the learnable parameters of the proposed framework, \(N\) is the total number of training pairs, \(x\) and \(y\) are the input and output image respectively, and \(MFNN(\cdot)\) is proposed framework for low-light image enhancement.
\textbf{Implementation:}
We train the proposed MFNN using Python (v3.8) coupled with PyTorch framework and conduct experiments on 
2x NVIDIA RTX 3090 GPUs with AMD Ryzen ThreadRipper 3960X CPU.   
We use 16 residual blocks split into three layers of the hierarchy, and generate 256 feature 
maps for each scale. We train the proposed MFNN on dataset provided in the challenge along with LoL Dataset \cite{wei2018deep} with patches of 
size 600*400. During training, we use Adam optimizer with $\beta_{1}=0.9$, $\beta_{2}=0.999$, 
$\epsilon=10e^{-8}$, and set the learning rate $lr=0.0002$.
We train the proposed MFNN for 1000 epochs. We optimize the learning with proposed $\mathcal{L}_{MFNN}$. During testing, we use full resolution images  ($2992\times2000$), on single RTX 3090 GPU. Average testing time for single image on full resolution is 0.53s on RTX 3090 GPU.

%% file: team17_221B/main.tex
\subsection{221B}
\begin{figure}[!htp]
    \centering
    \includegraphics[width=.9\linewidth]{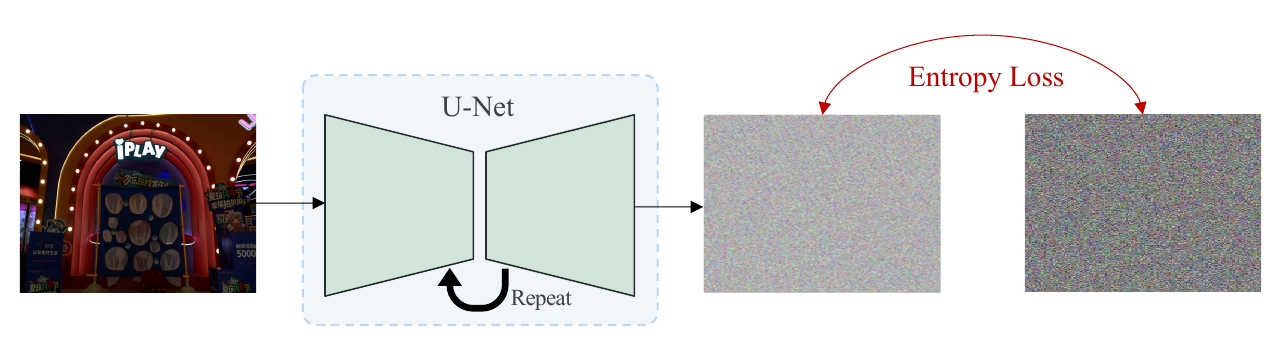}
    \caption{Overview of diffusion model training process with the differentiable spatial entropy loss proposed by the team 221B.}
    \label{fig:entropy-refusion}
\end{figure}
The organizer wants to highlight that this method has obtained the first place on the LPIPS ranking, with a score of 0.1084, outperforming the overall best-ranked work with a score of 0.1221.

\noindent\textbf{Description:} Most existing diffusion-based works~\cite{ho2020denoising} simply employ $\ell_1$/$\ell_2$ loss to train the noise prediction network in a deterministic way, which always produces smooth intermediate noises (or states) in the reverse process. Although the perceptual quality can be improved with longer diffusion steps, it is reasonable to explore a perceptual-based loss function facilitate the training process. In this work, they propose to learn the network from a pure statistic perspective, that is, \textit{directly learning distributions rather than pixel-by-pixel distances}. The core idea is the differentiable spatial entropy that drived from the kernel density estimation (KDE) and measures the spatial information of an image. Formally, following~\cite{parzen1962estimation,avi2023differentiable}, the two-dimensional KDE is defined as the probability of a specific intensity value $i$ and its neighborhood $j$ as:
\begin{equation}
    \begin{split}
    \centering
    \displaystyle \hat{f}_h(i, j) 
        &=\frac{2}{Nh} \sum_{x \in X} \frac{1}{2} \mathbb{I}\left \{ \frac{|x-i|}{h} < 1 \right \} \cdot \frac{1}{2}\mathbb{I}\left \{\frac{|\tilde{x}-j|}{h} < 1  \right \} \\
        &=\frac{2}{Nh } \sum_{x \in X} \mathcal{K}_1(\frac{x-i}{h}) \cdot \mathcal{K}_2(\frac{\tilde{x}-j}{h}),
    \end{split}
    \label{eq:kde2d_K}
\end{equation}
where $\mathcal{K}_1$ and $\mathcal{K}_2$ are arbitrary kernel functions (they can also be the same). To make use of the differentiable spatial KDE in image generation, this work considers to introduce the relative entropy (also called KL-divergence) to measure the distribution distance between the predicted image and ground truth. Let $P$ and $Q$ represent the ground truth's and prediction's probability distributions computed from~\cref{eq:kde2d_K}, the spatial relative entropy is defined to be
\begin{equation}
    \centering
    \displaystyle H_s(P, Q) = D_{KL}(P||Q) = -\sum_{i=0}^L \sum_{j \in \Omega} p_{i,j}\log \frac{q_{i,j}}{p_{i,j}}.
    \label{eq:skl}
\end{equation}
Their diffusion model is built on the mean-reverting SDE~\cite{luo2023image,luo2023refusion}. By optimizing the entropy-based distribution distance between predictions and ground truth states, their results have better visual qualities than that only using $\ell_1$/$\ell_2$ loss.~\cref{fig:entropy-refusion} illustrates the overview of the proposed diffusion training process.

\noindent\textbf{Implementation:} This team use the provided low-light dataset but further split 10 image pairs for validation. The Refusion framework is used as the baseline and the team directly changed its loss funcion to the proposed differentiable spatial entropy. In training, all images are cropped with patch size $128\times 128$, and the total training iteration is set to 500,000 with a batch 16. Following Refusion~\cite{luo2023refusion}, this work use the AdamW~\cite{loshchilov2017decoupled} optimizer and the `CosineDecay' scheduler with an initial learning rate 0.0001. And the diffusion steps are fixed to 100 in both training and testing.

%% file: team18_KLETech_CEVI_Dark_Knights/main.tex
\subsection{KLETech-CEVI\_Dark\_Knights}
\begin{figure}[!htp]
    \includegraphics[width=1\linewidth]{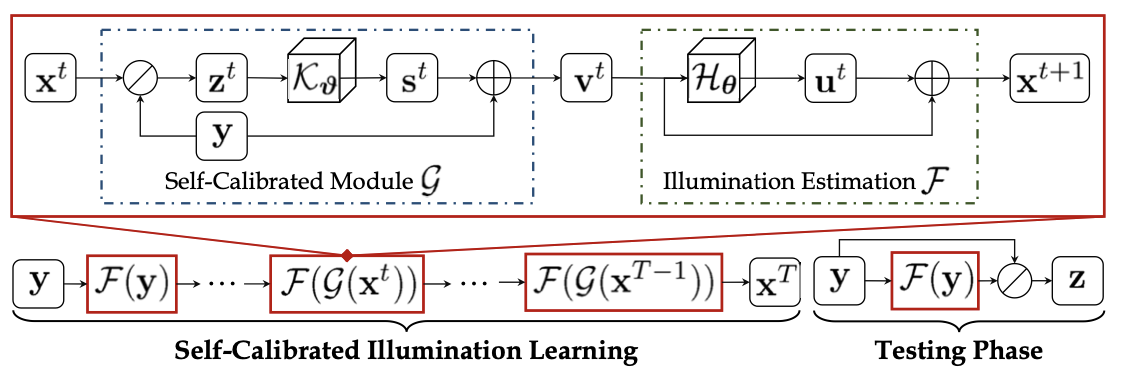}
    \caption{Framework of Self-Calibrated Illumination Module (Reproduced from \cite{ma2022toward}). We employ variant of this method for enhancement of low-light images with custom loss function.}
\end{figure}
\noindent\textbf{Description:}~In accordance with the Retinex theory, there exists a relationship between the observed low-light image, denoted as \( y \), and the desired clear image \( z \), expressed as \( y = z \otimes x \), where \( x \) signifies the illumination component. Typically, optimization efforts in low-light image enhancement primarily focus on refining the illumination component. To achieve enhanced output, one common approach involves removing the estimated illumination in accordance with the Retinex theory. Drawing inspiration from prior works on illumination optimization, particularly those employing a stage-wise approach, we propose a progressive perspective for modeling this task. Here, we introduce a mapping \( H_\theta \) parameterized by \( \theta \) to learn the illumination component, where each stage \( t \) (with \( t = 0, ..., T - 1 \)) consists of a basic unit represented as \( u_t = H_\theta(x_t) \) and \( x_{t+1} = x_t + u_t \), with \( x_0 = y \). It's worth noting that we adopt a weight sharing mechanism across stages, utilizing the same architecture \( H \) and weights \( \theta \) for each stage. Specifically, the parameterized operator \( H_\theta \) effectively learns a simple residual representation \( u_t \) between the illumination and the observed low-light image. This process is motivated by the consensus that illumination and low-light observations exhibit similarities or linear connections in most areas.

We establish a module to ensure the convergence of results from each stage to a consistent state. Given that the input of each stage originates from the previous stage, and the initial input of the first stage is predetermined as the low-light observation, we propose to indirectly explore convergence behavior by bridging the input of each stage (excluding the first) with the low-light observation. To achieve this, we employ a self-calibrated module \cite{ma2022toward} denoted as \( s \), which is added to the low-light observation to represent the disparity between the input of each stage and the initial stage. Specifically, the self-calibrated module, represented as \( s = K_\theta(z) \), utilizes the parameterized operator \( K_\theta \) with learnable parameters \( \theta \) to generate \( v = y + s \), where \( v_t \) denotes the converted input for each stage (\( t \geq 1 \)). Consequently, the conversion for the basic unit in the \( t^{th} \) stage (\( t \geq 1 \)) can be expressed as \( F(x_t) \rightarrow F(G(x_t)) \). This self-calibrated module effectively corrects the input of each stage by integrating physical principles, thereby indirectly influencing the output of each stage. 
In addition to this, we employ NIQE (Natural Image Quality Evaluator) \cite{mittal2012making} as a loss function. NIQE offers a valuable approach for improving perceptual quality. By utilizing NIQE as a metric to quantify the naturalness and perceptual fidelity of enhanced images, the training process aims to minimize the NIQE scores of the generated images. This strategy encourages the model to produce output images that closely resemble natural, high-quality images, enhancing their visual appeal and usability. By optimizing for lower NIQE scores, the model learns to prioritize perceptually pleasing enhancements, ultimately leading to outputs that are not only visually appealing but also retain important natural characteristics. Integrating NIQE as a loss function thus contributes to the development of low-light image enhancement techniques that align closely with human perception and preferences. NIQE as a loss function is given as:
\begin{equation}
L_{\text{NIQE}} = \frac{1}{N} \sum_{i=1}^{N} \text{NIQE}(F(G(x_{t,i}))),
\end{equation}
where \(N\) is the number of samples, \(x_{t,i}\) represents the input to the \(t\)-th stage for the \(i\)-th sample, \(F(G(x_{t,i}))\) denotes the output of the \(t\)-th stage for the \(i\)-th sample after applying the self-calibrated module, and \(\text{NIQE}\) is the function that calculates the NIQE score for an image.
Minimizing this loss function during training would encourage the model to produce output images with lower NIQE scores, indicating higher perceptual quality and naturalness. This would result in output images that are closer in quality to natural, high-quality images.\\
\textbf{Implementation:}
We train the proposed method using Python (v3.8) coupled with PyTorch framework and conduct experiments on 
2x NVIDIA RTX 3090 GPUs with AMD Ryzen ThreadRipper 3960X CPU.   
We use 16 residual blocks split into three layers of the hierarchy, and generate 256 feature 
maps for each scale. We train the proposed method on dataset provided in the challenge along with LoL Dataset \cite{wei2018deep} with images of 
size 2992*2000. During training, we use Adam optimizer with $\beta_{1}=0.9$, $\beta_{2}=0.999$, 
$\epsilon=10e^{-8}$, and set the learning rate $lr=0.0002$.
We train the proposed method for 1000 epochs.
During testing, we use full resolution images  (2992 * 2000), on single RTX 3090 GPU. Average testing time for single image on full resolution is 0.53s on RTX 3090 GPU.

%% file: team19_BFU_LL/main.tex
\subsection{BFU-LL}
\begin{figure}[!htp]
  \centering
   \includegraphics[width=\linewidth]{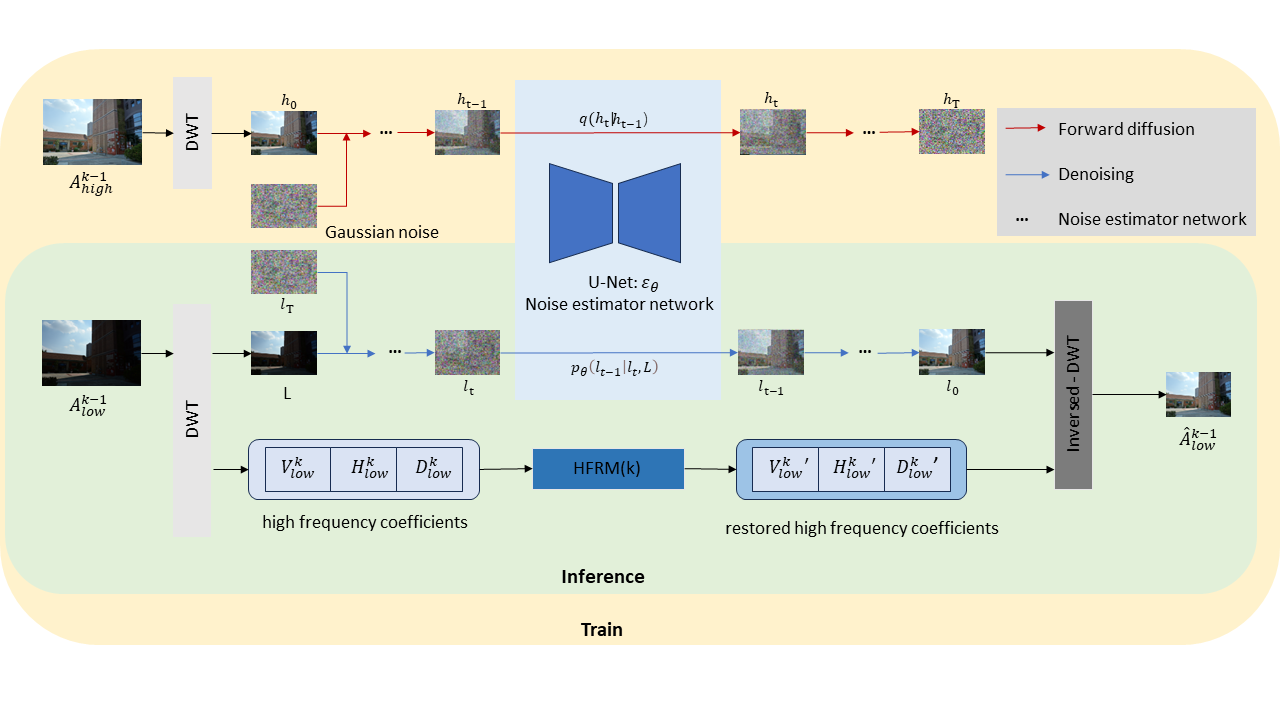}
   \caption{The network architecture of team BFU-LL.}
   \label{bfu_DiffLL}
\end{figure}
\noindent\textbf{Description:}
Diverging from the use of more complex, Transformer-based models for image restoration \cite{self_att1,self_att2}, this team focuses on addressing low-light issues through the application of diffusion methods. They introduce a novel approach for enhancing low-light images proposed by \cite{lowlight_diffusion}, as illustrated in~\cref{bfu_DiffLL}. The method first converting these images into the wavelet domain via 2D discrete wavelet transformation (DWT), iterated multiple times. This process yields an average coefficient and sets of high-frequency coefficients. By employing Haar wavelets \cite{2d_haar_dwt}, the method divides the input into four sub-bands, effectively capturing global information and sparse local details. This significantly reduces spatial dimensions while preserving essential image information. The team's wavelet-based diffusion model utilizes a forward diffusion process during the training phase and a denoising process across both training and inference phases. Moreover, they have developed a High-Frequency Restoration Module (HFRM) for the reconstruction of high-frequency details. Utilizing the U-Net architecture as the noise estimator network and implementing a fixed variance schedule, their method methodically transforms the input into corrupted noise data, subsequently employing Gaussian denoising transitions to achieve a clear and enhanced output.

Building on the methodologies described in previous studies \cite{hybrid_loss1, hybrid_loss2}, this team design and adopt a hybrid loss function as follows:

\begin{equation}
\mathcal{L}_{\text{total}} = \mathcal{L}_{\text{smooth}} + \alpha \mathcal{L}_{\text{MS-SSIM}} + \beta \mathcal{L}_{\text{per}},
\end{equation}
where $\alpha = 0.01$, $\beta = 0.01$, and $\gamma = 0.005$ are the hyperparameters weighting each loss component.

\noindent\textbf{Implementation:} The implementation by this team is conducted using PyTorch. The proposed network effectively converges after undergoing training for $1 \times 10^{4}$ iterations on a system equipped with four NVIDIA RTX 1080Ti GPUs. For optimization, the Adam optimizer is employed. Considering the impact of the learning rate on experimental results \cite{qlabgrad}, the initial learning rate is set to $1 \times 10^{-4}$. This learning rate decays by a factor of 0.8 after every $5 \times 10^{3}$ iterations. The team utilizes a batch size of 12 and a patch size of $256 \times 256$. The scale $K$ for the wavelet transformation is set to 2.

%% file: team20_SVNIT_NTNU/main.tex
\subsection{SVNIT\_NTNU}
\begin{figure}[!htp]
    \centering
    \subfloat[The block schematic of the proposed architecture for Low Light Enhancement.]{\includegraphics[width=1.0\linewidth]{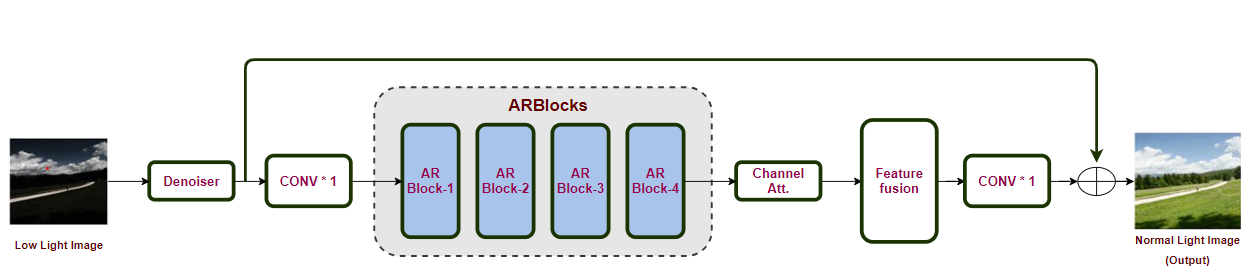}}\\
    \subfloat[The design of the ARblock used in the proposed model. ]{\includegraphics[width=1.0\linewidth]{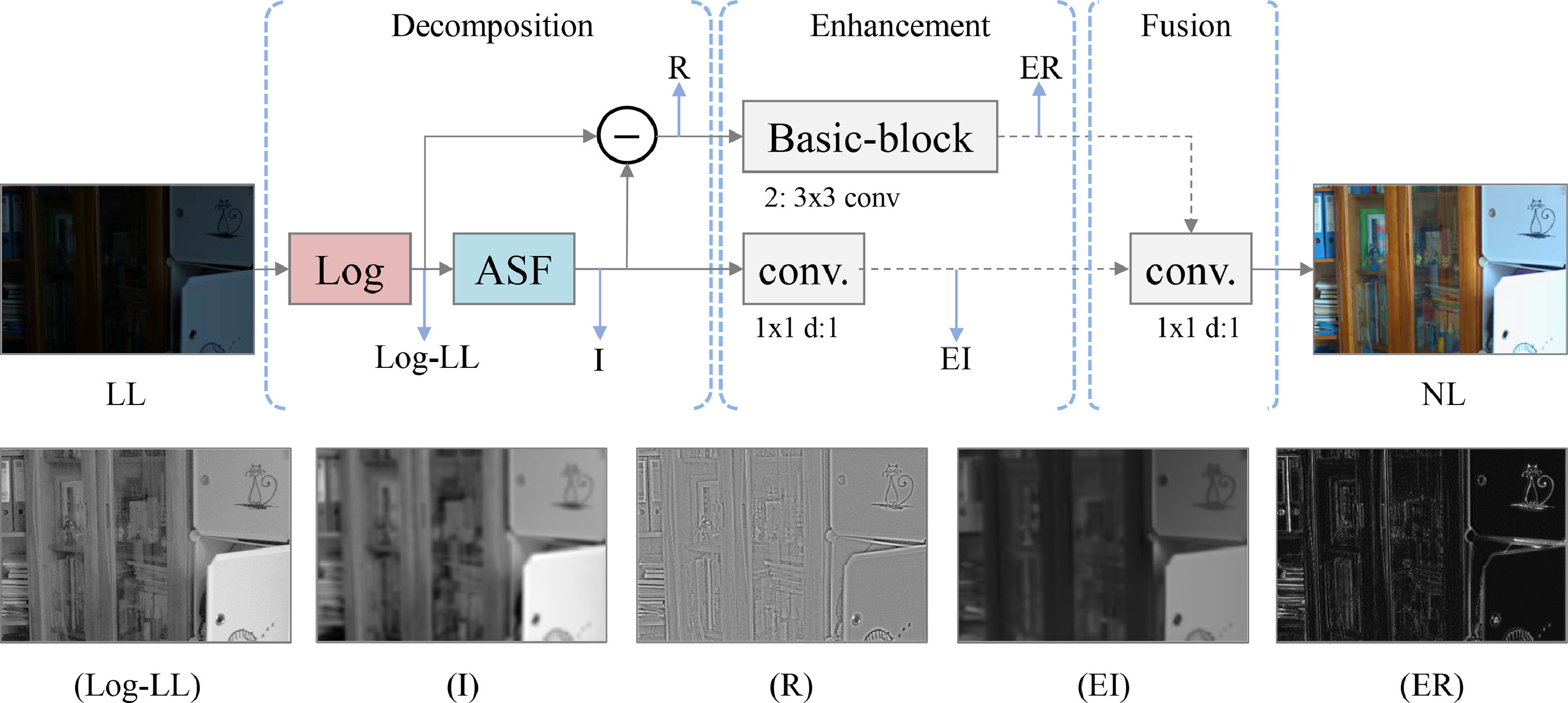}}
    \caption{The architecture of the proposed model for Low Light Enhancement.}
    \label{fig:team20_arch}
\end{figure}
\noindent\textbf{Description:}
In order to design low light image enhancement,we use multiple Retinex blocks and multiple channel attention in the proposed solution. The~\cref{fig:team20_arch}(a) depicts the proposed architecture for Low Light enhancement. The initial step involves feeding the low-light image into a denoiser module tailored to address the noise typical of low-exposure conditions. Subsequently, a shallow feature extraction process, achieved through a single convolutional operation, projects the denoised image into a feature space.

Following this, the image undergoes enhancement through multi-branch Adaptive Retinex (AR) Blocks operating at various scales. These blocks simultaneously enhance both broad and fine details within the image.

The shallow features and the enhanced results are then merged within a multiple channel attention module, which effectively extracts global and local features and combines them using a sigmoid function.

Lastly, the fused features are refined through another convolutional layer, culminating in the production of a clearer, brighter rendition of the original low-light image. Despite the addition of the channel attention module, which marginally increases parameter count, the overall network remains lightweight, owing to the efficiency achieved through the AR Blocks and denoiser modules.


The network incorporates multiple Adaptive Retinex Blocks (ARBlocks), which represent a novel extension of the Single Scale Retinex method in feature space~\cite{retinexformer}. At the core of each ARBlock lies an efficient illumination estimation function known as the Adaptive Surround Function (ASF) \cite{lowlight}. This function, akin to a general form of surround functions, is implemented using convolutional layers. The ARBlock serves a dual purpose, facilitating both illumination adjustment and reflectance enhancement within the image.~\cref{fig:team20_arch} illustrates the architecture of the ARBlock, demonstrating its capability to enhance both low-frequency and high-frequency components of the image, respectively.

\noindent\textbf{Implementation:~}The code is implemented using Pytorch library. The proposed network is trained using weighted combination of $l_1$, SSIM loss, and charbornnier loss with a learning rate of $1 \times 10^{-4}$ which is decayed by $1\times10^4$ iterations and the same is optimized using Adam optimizer. The model is trained up to $1 \times 10^5$ iterations with a batch size of $4$.

%% file: team21_yanhailong/main.tex
\subsection{yanhailong}
\begin{figure}  
	\centering
\includegraphics[width=1\linewidth]{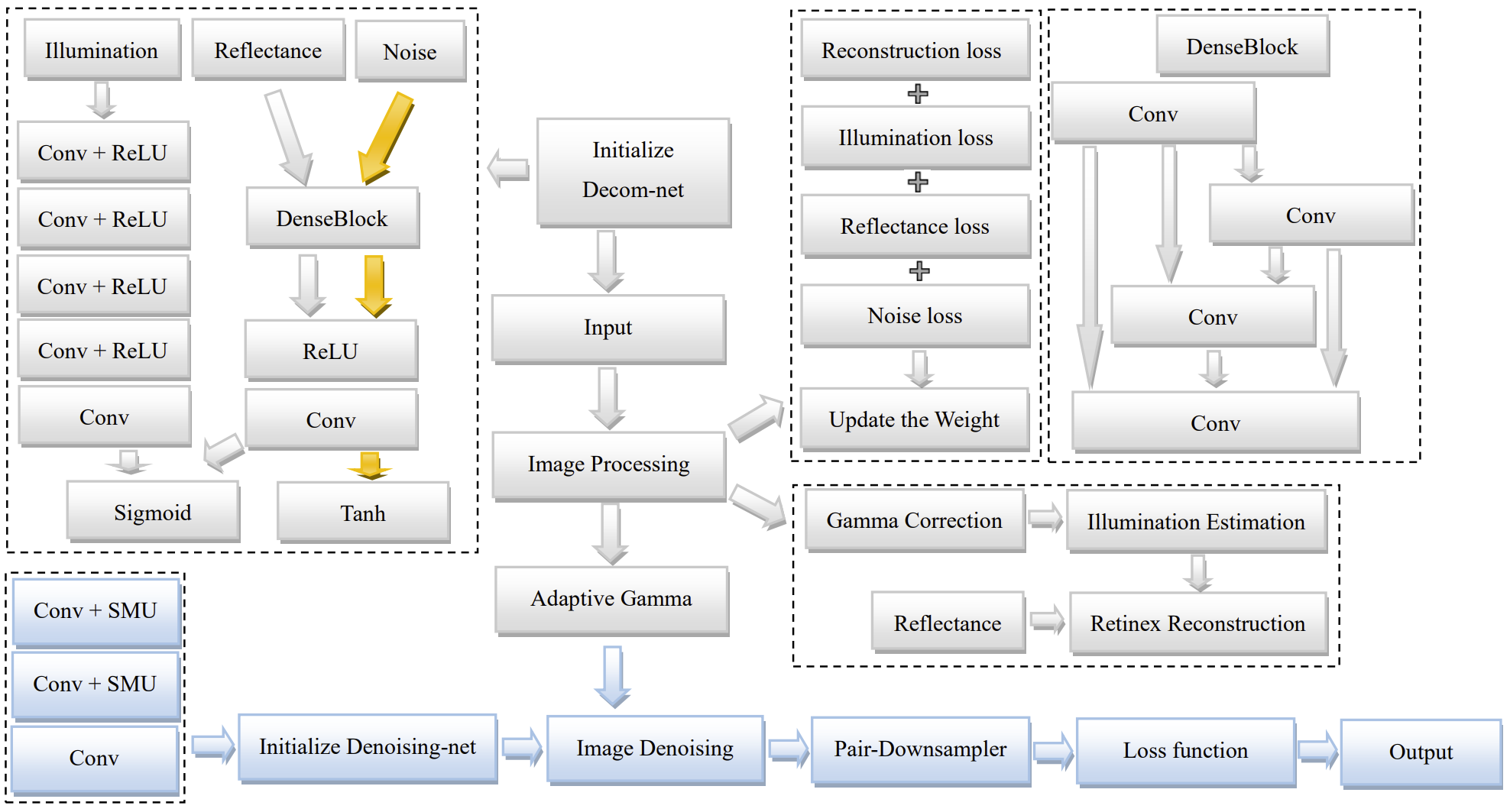}
	\caption{General method flow chart.}
	\label{yanhailong}
\end{figure}
\noindent\textbf{Description:}~We propose ZSRADNet:~Zero-Shot Retinex Joint Adaptive-Illumination and Denoising for Underexposed Image Enhancement, which mainly considers underexposed images often suffer severe quality degradation. Many scenes exhibit issues related to occlusion and shadows, resulting in non-uniform illumination in the images. Most methods for restoring underexposed images rely on global enhancement, which can lead to excessive enhancement in areas that were originally well-lit. Both supervised and unsupervised learning approaches have limitations: either poor generalization or susceptibility to unstable training. To address these challenges, the proposed a Zero-Shot method that learns enhancement solely from test images. It consists of two main components: The Decomposition Network and The Denoising Network.

As shown in~\cref{yanhailong}, inspired by \cite{zhu2020zero}, the Decomposition network employs a dual-branch structure to break down the input image into three components: Illumination, Reflection, and Noise. Initially, the input image passes through a simple four-layer CNN (Convolutional Neural Network) combined with ReLU (Rectified Linear Unit) activation and a Sigmoid function to obtain the illumination. Next, a designed DenseBlock layer, combined with ReLU and Convolution, generates the reflection (using Sigmoid activation) and the noise (using Tanh activation). DenseBlock can make better use of the feature information of the previous layer and help to improve feature reusability.

To achieve Zero-Shot enhancement, the proposed combines loss functions of Retinex reconstruction, Reflectance smoothness, and Illumination-guided noise estimation. By iteratively minimizing this combined loss function, the proposed effectively estimates the noise and restores the illumination. Additionally, the proposed introduces an adaptive gamma correction method that uses illumination as a measure of non-uniformly lit regions in the image. It brightens underexposed areas while leaving well-lit regions untouched. Our loss function consists of three parts:
\begin{equation}\label{yan(1)}
 L= L_{recon}+{\lambda_{ref}}\cdot{L_{ref}}+{\lambda_{noise}}\cdot{L_{noise}},
\end{equation}
where $L_{recon}$, $L_{ref}$ and $L_{noise}$ represent reconstruction loss, reflectance smoothness loss, and Illumination-guided noise loss respectively, and $\lambda_{ref}$, $\lambda_{noise}$ represents corresponding weight factors.





The Denoising network generates a pair of noise maps \cite{mansour2023zero} from a single noise image and utilizes these maps to train a simple three-layer network (CNN combined with SMU activation \cite{biswas2022smooth}). Specifically, the proposed begins by applying a down-sampling operator to decompose the output from the Decomposition network into a pair of down-sampled images. One of these observations serves as the input value, while the other acts as the target. The proposed then employs regularization techniques during training to achieve effective denoising.

As a method specific to a single input image, it does not require any prior image samples or prior training models. The final enhanced results have advantages in the brightness and naturalness of non-uniform illumination images.

\noindent\textbf{Implementation:}~The team uses Pytorch and the test device is the A6000 GPU.
The illumination branch of the Decomposition network comprises a four-layer structure consisting of $3\times3$ Convolution followed by ReLU activation functions. On the other hand, the reflection and noise branches utilize a four-layer architecture of $3\times3$ Convolution with Denseblock, followed by ReLU activation functions, culminating in an additional layer of $3\times3$ Convolution. The network undergoes a total of $1000$ iterations during training. In the Denoising network, two layers of $3\times3$ Convolution are employed, each followed by SMU activation, and ultimately connected to a final layer of $3\times3$ Convolution. The training process utilizes Mean Squared Error (MSE) loss is used for $2000$ iterations. All the learning rates were $0.01$; the optimizer uses Adam; without training, the input of a single image directly outputs the enhanced result.

%% file: team22_Mishka/main.tex
\subsection{Mishka}
\begin{figure}[ht]
    \centering
    \includegraphics[width=1.0\linewidth]{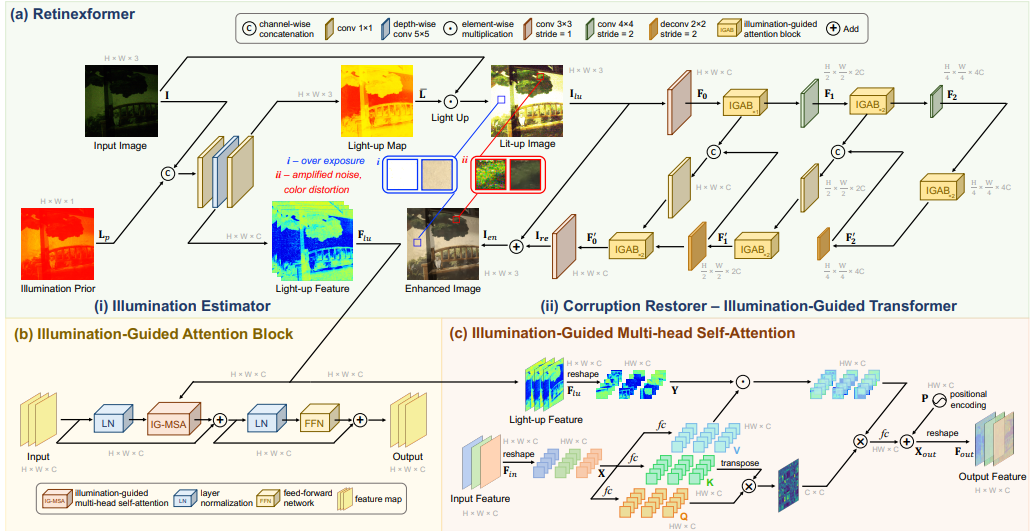} 
    \caption{Illustrative diagram showing the application of Retinexformer for low-light enhancement.}
    \label{fig:team22_diagram}
\end{figure}
\noindent\textbf{Description:~}As shown in~\cref{fig:team22_diagram}, we employed the Retinexformer~\cite{cai2023retinexformer} model, a one-stage Retinex-based transformer, specifically designed for low-light image enhancement tasks. Our approach is distinguished by training the Retinexformer model directly on the dataset provided by the competition, demonstrating its adaptability and generalization capability across different datasets. This direct training method underscores the model's effectiveness and flexibility for specific low-light conditions. Our method maintains a moderate level of complexity throughout all stages, ensuring efficiency and effectiveness from data preprocessing to model training and final evaluation.

\noindent\textbf{Implementation:~}Our training configuration details various aspects of the model training process. We conducted a total of 300,000 iterations without a warm-up phase, and gradient clipping was employed to prevent gradient explosion.
During training, we divided the 300,000 iterations into two cycles:
\begin{itemize}
  \item The first cycle involved a fixed learning rate of 3$e$-4 for the initial 92,000 iterations.
  \item The second cycle applied a cosine annealing strategy, decreasing the learning rate from 3$e$-4 to 1$e$-6 over the subsequent 208,000 iterations.
\end{itemize}

We also utilized mixed data augmentation techniques , and for the optimizer configuration, we chose the Adam optimizer with an initial learning rate of 2$e$-4 and betas set to [0.9, 0.999]. Regarding the loss function, we employed the $L_1$ Loss, with a loss weight of 1 and reduction method set to mean.

%% file: team01_SYSU_FVL_T2/affiliation.tex
\subsection*{SYSU-FVL-T2}
\noindent\textit{\textbf{Title: }}Scale-Robust Low-light Ultra-High-Definition Image Enhancement\\
\noindent\textit{\textbf{Members: }} \\
Zhi Jin$^{1,2}$ (\href{mailto:jinzh26@mail.sysu.edu.cn}{jinzh26@mail.sysu.edu.cn}),
Hongjun Wu$^1$, 
Chenxi Wang$^1$, 
Haitao Ling$^1$ \\
\noindent\textit{\textbf{Affiliations: }} \\ 
$^1$ School of Intelligent Systems Engineering, Shenzhen Campus of Sun Yat-sen University, Shenzhen, Guangdong 518107, China \\
$^2$ Guangdong Provincial Key Laboratory of Fire Science and Technology, Guangzhou 510006, China\\

%% file: team02_Retinexformer/affiliation.tex
\subsection*{Retinexformer}
\noindent\textit{\textbf{Title: }} Retinexformer: One-stage Retinex-based Transformer for Low-light Image Enhancement\\
\noindent\textit{\textbf{Members: }} \\
Yuanhao Cai$^1$ (\href{mailto:caiyuanhao1998@gmail.com}{caiyuanhao1998@gmail.com}),
Hao Bian$^2$,
Yuxin Zheng$^2$,
Jing Lin$^2$,
Alan Yuille$^1$ \\
\noindent\textit{\textbf{Affiliations: }} \\ 
$^1$ Johns Hopkins University \\
$^2$ Tsinghua University \\

%% file: team03_DH_AISP/affiliation.tex
\subsection*{DH-AISP}
\noindent\textit{\textbf{Title: }} Multi-scale feature fusion Low-light Image Enhancement\\
\noindent\textit{\textbf{Members: }} \\
Ben Shao$^1$ (\href{mailto:2830802483@qq.com}{2830802483@qq.com}),
Jin Guo$^1$, 
Tianli Liu$^1$,
Mohao Wu$^1$ \\
\noindent\textit{\textbf{Affiliations: }} \\ 
$^1$ Zhejiang Dahua Technology Co.,Ltd. \\

%% file: team04_NWPU_DiffLight/affiliation.tex
\subsection*{NWPU-DiffLight}
\noindent\textit{\textbf{Title: }} DiffLight: Integrating Content and Detail for Low-light Image Enhancement\\
\noindent\textit{\textbf{Members: }} \\
Yixu Feng$^1$ (\href{yixu-nwpu@mail.nwpu.edu.cn}{yixu-nwpu@mail.nwpu.edu.cn}),
Shuo Hou$^1$,
Haotian Lin$^2$,
Yu Zhu$^1$,
Peng Wu$^1$,
Wei Dong$^3$,
Jinqiu Sun$^1$,
Yanning Zhang$^1$,
Qingsen Yan$^1$ \\
\noindent\textit{\textbf{Affiliations: }} \\ 
$^1$ School of Computer Science, Northwestern Polytechnical University, China \\
$^2$ School of Software, Northwestern Polytechnical University, China
$^3$ School of Computer Science, Xi’an University of Architecture and Technology, China \\

%% file: team05_GiantPandaCV/affiliation.tex
\subsection*{GiantPandaCV}
\noindent\textit{\textbf{Title: }} UHDMamba: Towards Effective and Efficient State-Space Model for UHD Low-light Image Enhancement\\
\noindent\textit{\textbf{Members: }} \\
Wenbin Zou$^1$ (\href{mailto:alexzou14@foxmail.com}{alexzou14@foxmail.com}),
Weipeng Yang$^1$, 
Yunxiang Li$^2$, 
Qiaomu Wei$^3$,
Tian Ye$^4$, 
Sixiang Chen$^{3,4}$ \\
\noindent\textit{\textbf{Affiliations: }} \\ 
$^1$ South China University of Technology, China \\
$^2$ Fuzhou University, China\\
$^3$ Chengdu University of Information Technology, China \\
$^4$ Hong Kong University of Science and Technology (Guangzhou), China \\

%% file: team06_LVGroup_HFUT/affiliation.tex
\subsection*{LVGroup\_HFUT}
\noindent\textit{\textbf{Title: }} Exploring the Application of NAFNet in Low Light Enhancement\\
\noindent\textit{\textbf{Members: }} \\
Zhao Zhang$^1$ (\href{mailto: cszzhang@gmail.com}{cszzhang@gmail.com}),
Suiyi Zhao$^1$, 
Bo Wang$^1$,
Yan Luo$^1$,
Zhichao Zuo$^1$,
Mingshen Wang$^1$,
Junhu Wang$^1$,
Yanyan Wei$^1$ \\
\noindent\textit{\textbf{Affiliations: }} \\ 
$^1$ Hefei University of Technology, China\\

%% file: team07_Try1try8/affiliation.tex
\subsection*{Try1try8}
\noindent\textit{\textbf{Title: }} Low Enhancement Model Ensemble of Multi-patch RetinexFormers\\
\noindent\textit{\textbf{Members: }} \\
Xiaopeng Sun$^1$ (\href{mailto:xpsun@stu.xidian.edu.cn}{xpsun@stu.xidian.edu.cn}), 
Yu Gao$^1$ ,
Jiancheng Huang$^1$  \\
\noindent\textit{\textbf{Affiliations: }} \\ 
$^1$ Individual Researcher \\

%% file: team08_Pixel_warrior/affiliation.tex
\subsection*{Pixel\_warrior}
\noindent\textit{\textbf{Title: }} MLP for Low-Light Image Enhancement\\
\noindent\textit{\textbf{members:~}} \\
Hongming Chen$^1$ (\href{mailto:chenhongming@stu.sau.edu.cn}{chenhongming@stu.sau.edu.cn}),
Xiang Chen$^2$ \\
\noindent\textit{\textbf{Affiliations: }} \\ 
$^1$ Shenyang Aerospace University, China\\
$^2$ Nanjing University of Science and Technology, China\\

%% file: team09_HuiT/affiliation.tex
\subsection*{HuiT}
\noindent\textit{\textbf{Title:~}} Enhanced Low-Light Imaging Method Based on Axis-Transformer\\
\noindent\textit{\textbf{Members: }} \\
Hui Tang$^1$ (\href{mailto:yourEmail@xxx.xxx}{2639442956@qq.com}),
Yuanbin Chen$^1$,
Yuanbo Zhou$^1$,
Xinwei Dai$^1$,
Xintao Qiu$^1$,
Wei Deng$^2$, 
Qinquan Gao$^{1,2}$,
Tong Tong$^{1,2}$ \\
\noindent\textit{\textbf{Affiliations: }} \\ 
$^1$ Fuzhou University, Fuzhou, China \\
$^2$ Imperial Vision Technology, Fuzhou, China \\

%% file: team10_X_LIME/affiliation.tex
\subsection*{X-LIME}
\noindent\textit{\textbf{Title:~}}
Supervised Deep Curve Estimation with Denoiser
\noindent\textit{\textbf{Members:~}} \\
Mingjia Li$^1$ (\href{mailto:mingjiali@tju.edu.cn}{mingjiali@tju.edu.cn}),
Jin Hu$^1$,
Xinyu He$^1$,
Xiaojie Guo$^1$ \\
\noindent\textit{\textbf{Affiliations: }} \\ 
$^1$ Tianjin University, China \\

%% file: team11_Image_Lab/affiliation.tex
\subsection*{Image Lab}
\noindent\textit{\textbf{Title: }} Enhancing Low-Light Images: A SCPA-Based Approach with RDCA-UNet Architecture\\
\noindent\textit{\textbf{Members: }} \\
Sabarinathan$^1$  (\href{mailto:sabarinathantce@gmail.com}{sabarinathantce@gmail.com}),
K Uma$^2$,
A Sasithradevi$^3$,
B Sathya Bama$^4$,
S. Mohamed Mansoor Roomi$^4$,
V.Srivatsav$^5$ \\
\noindent\textit{\textbf{Affiliations: }} \\ 
$^1$ Couger Inc,Japan \\
$^2$ Sasi Institute of Technology \& Engineering, India\\
$^3$ Vellore Institute of Technology, India\\
$^4$ Thiagarajar college of engineering, India \\
$^5$ Coventry University, United Kingdom \\

%% file: team12_dgzzqteam/affiliation.tex
\subsection*{dgzzqteam}
\noindent\textit{\textbf{Title: }} One-stage gatebased
Framework\\
\noindent\textit{\textbf{Members: }} \\
Jinjuan Wang (\href{mailto:yourEmail@xxx.xxx}{15600951607@163.com})$^1$, Long Sun$^1$, Qiuying Chen$^1$, Jiahong Shao$^1$, Yizhi Zhang$^1$
\\
\noindent\textit{\textbf{Affiliations: }} \\ 
$^1$ Independent Researchers \\

%% file: team13_Cidaut_AI/affiliation.tex
\subsection*{Cidaut AI}
\noindent\textit{\textbf{Title: }} \textbf{InstructIR}: High-Quality Image Restoration Following Human Instructions \\
\url{https://github.com/mv-lab/InstructIR} \\
\noindent\textit{\textbf{Members:}} \\
Marcos V. Conde$^{1,2}$ (\href{mailto:marcos.conde@uni-wuerzburg.de}{marcos.conde@uni-wuerzburg.de}),
Daniel Feijoo$^1$, 
Juan C. Benito$^1$, 
Alvaro Garc\'{i}a$^1$ \\
\noindent\textit{\textbf{Affiliations: }} \\ 
$^1$ Cidaut AI \\
$^2$ CVLab, University of Wuerzburg \\

%% file: team14_OptDev/affiliation.tex
\subsection*{OptDev}
\noindent\textit{\textbf{Title: }} Learning Optimized Low-Light Image Enhancement for Edge Vision Tasks\\
\noindent\textit{\textbf{Members: }} \\
Jaeho Lee$^1$  (\href{mailto:jaeho.lee@opt-ai.kr}{jaeho.lee@opt-ai.kr}), Seongwan Kim$^1$, Sharif S M A$^1$, Nodirkhuja Khujaev$^1$, Roman Tsoy$^1$\\
\noindent\textit{\textbf{Affiliations: }} \\ 
$^1$ Opt-AI 

%% file: team15_ataza/affiliation.tex
\subsection*{ataza}
\noindent\textit{\textbf{Title: }} An Efficient Frequency Guided Image Enhancement Network for Low Light
Image Enhancement and Shadow Removal\\
\noindent\textit{\textbf{Members: }} \\
Ali Murtaza$^{1,2}$ (\href{mailto:ali.murtaza.ali29@outlook.com}{ali.murtaza.ali29@outlook.com}),\\
Uswah Khairuddin$^{1,2}$, 
Ahmad 'Athif Mohd Faudzi$^{2}$ \\
\noindent\textit{\textbf{Affiliations: }} \\ 
$^1$ Malaysia-Japan International Institute of Technology (MJIIT), University Teknologi Malaysia, Kuala Lumpur, , Malaysia \\
$^2$ Center for Artificial Intelligence and Robotics (CAIRO), Universiti Teknologi Malaysia, Kuala Lumpur, Malaysia \\

%% file: team16_KLETech_CEVI_LowlightHypnotise/affiliation.tex
\subsection*{KLETech-CEVI\_LowlightHypnotise}
\noindent\textit{\textbf{Title:~}}Hierarchical LightNet: Towards Low-Light Image Enhancement\\ 
\noindent\textit{\textbf{Members:~}} \\ 
Sampada Malagi$^{1,3}$ (\href{mailto:sampadamalagi12@gmail.com}{sampadamalagi12@gmail.com}),
Amogh Joshi$^{1}$,
Nikhil Akalwadi$^{1,3}$,
Chaitra Desai$^{1,3}$,
Ramesh Ashok Tabib$^{1,2}$,
Uma Mudenagudi$^{1,2}$ \\
\noindent\textit{\textbf{Affiliations: }}\\
$^1$ Center of Excellence in Visual Intelligence (CEVI), KLE Technological University, Hubballi, Karnataka, India\\ 
$^2$ School of Electronics and Communication Engineering, KLE Technological University, Hubballi, Karnataka, India\\ 
$^3$ School of Computer Science and Engineering, KLE Technological University, Hubballi, Karnataka, India\\

%% file: team17_221B/affiliation.tex
\subsection*{221B}
\noindent\textit{\textbf{Title: }} Equipping Diffusion Models with Differentiable Spatial Entropy for Low-Light Image Enhancement\\
\noindent\textit{\textbf{Members: }} \\
Wenyi Lian$^1$ (\href{mailto:shermanlian@163.com}{shermanlian@163.com}),
Wenjing Lian$^2$ \\
\noindent\textit{\textbf{Affiliations: }} \\ 
$^1$ Uppsala University \\
$^2$ Northeastern University \\

%% file: team18_KLETech_CEVI_Dark_Knights/affiliation.tex
\subsection*{KLETech-CEVI\_Dark\_Knights}
\noindent\textit{\textbf{Title: }}KnightNet: Illumination Guided Low Light Image Enhancement Network
\noindent\textit{\textbf{Members: }} \\ 
Jagadeesh Kalyanshetti$^{1,3}$ (\href{mailto:01fe21bcs004@kletech.ac.in}{01fe21bcs004@kletech.ac.in}),
Vijayalaxmi Ashok Aralikatti$^{1,3}$,
Palani Yashaswini$^{1,2}$,
Nitish Upasi$^{1}$,
Dikshit Hegde$^{1}$,
Ujwala Patil$^{1,2}$,
Sujata C$^{1,3}$ \\
\noindent\textit{\textbf{Affiliations: }}\\
$^1$ Center of Excellence in Visual Intelligence (CEVI), KLE Technological University, Hubballi, Karnataka, India\\ 
$^2$ School of Electronics and Communication Engineering, KLE Technological University, Hubballi, Karnataka, India\\ 
$^3$ School of Computer Science and Engineering, KLE Technological University, Hubballi, Karnataka, India\\

%% file: team19_BFU_LL/affiliation.tex
\subsection*{BFU-LL}
\noindent\textit{\textbf{Title: }} A Diffusion Model for Low-Light Enhancement\\
\noindent\textit{\textbf{Members: }} \\
Xingzhuo Yan$^1$ (\href{ayx1sgh@bosch.com}{ayx1sgh@bosch.com}),
Wei Hao$^2$,
Minghan Fu$^3$ \\
\noindent\textit{\textbf{Affiliations: }} \\ 
$^1$ Bosch Investment Ltd. \\
$^2$ Fortinet, Inc. \\
$^3$ University of Saskatchewan \\

%% file: team20_SVNIT_NTNU/affiliation.tex
\subsection*{SVNIT\_NTNU}
\noindent\textit{\textbf{Title:}}  Retinex based Low Light Image Enhancement\\
\noindent\textit{\textbf{Members: }} \\
Pooja choksy$^1$ (\href{mailto:yourEmail@xxx.xxx}{ds22ec003@eced.svnit.ac.in}),
Anjali Sarvaiya$^1$,
Kishor Upla$^1$,
Kiran Raja$^2$ \\

\noindent\textit{\textbf{Affiliations: }} \\ 
$^1$ Sardar Vallabhbhai National Institute of Technology, India \\
$^2$ Norwegian University of Science and Technology, Norway \\

%% file: team21_yanhailong/affiliation.tex
\subsection*{yanhailong}
\noindent\textit{\textbf{Title:}} ZSRADNet: Zero-Shot Retinex Joint Adaptive-Illumination and Denoising for Underexposed Image Enhancement\\
\noindent\textit{\textbf{Members: }} \\
Hailong Yan (\href{mailto:yhl00825@163.com}{yhl00825@163.com}) \\
\noindent\textit{\textbf{Affiliations: }} \\ 
University of Electronic Science and Technology of
China, China \\ 

%% file: team22_Mishka/affiliation.tex
\subsection*{Mishka}
\noindent\textit{\textbf{Title: }}Application and Adaptation of Retinexformer
\noindent\textit{\textbf{Members: }} \\
Yunkai Zhang$^1$ (\href{mailto:yourEmail@xxx.xxx}{1585832651@qq.com}),
Baiang Li$^1$, 
 Jingyi Zhang$^1$,
 Huan Zheng\(^2\) \\
\noindent\textit{\textbf{Affiliations: }} \\ 
$^1$ Hefei University of Technology, China \\
$^2$ University of Macau, China\\